\documentclass[final,3p,times]{elsarticle}

\usepackage{bm}
\usepackage{float}
\usepackage{amssymb}
\usepackage{amsthm}
\usepackage{amsmath}
\usepackage{multirow}
\usepackage{textcomp}
\usepackage{subcaption} 
\usepackage{siunitx}
\usepackage{epstopdf}
\usepackage[table]{xcolor}
\usepackage{xcolor}
\usepackage{colortbl}
\usepackage{graphicx}  
\usepackage{caption}   
\usepackage[utf8]{inputenc}
\usepackage[skip=4pt]{caption}
\journal{Pattern Recognition}

\begin{document}

\begin{frontmatter}



\title{FMRFusion: Frequency-Aware Multi-View Representation Learning for Heterogeneous Image Fusion}


\author[sdu,SDu]{Tao Zhou}
\ead{202520858@mail.sdu.edu.cn}
\affiliation[sdu]{organization={Control Science and Engineering, Shandong University},
            city={Jinan},
            postcode={250061},
            country={China}}
            
\affiliation[SDu]{organization={Key Laboratory of Machine Intelligence and System Control, Ministry of Education},
            city={Jinan},
            postcode={250061},
            country={China}}

\author[sdu,SDu]{Yunlong Liu}
\ead{202520833@mail.sdu.edu.cn}

\author[sdu,SDu]{Qinghui Chen}
\ead{202420785@mail.sdu.edu.cn}
            
\cortext[cor1]{Corresponding author}

\author[sdu,SDu]{Zekai Zhang}
\ead{202420810@mail.sdu.edu.cn}

\author[sdu,SDu]{Minlong Sun}
\ead{202535007@mail.sdu.edu.cn}

\author[sdu,SDu]{Changlin Bian}
\ead{202534975@mail.sdu.edu.cn}

\author[Macau]{Dagang Li}
\ead{dgli@must.edu.mo}
\affiliation[Macau]{organization={School of Computer Science and Engineering, Macau University of Science and Technology},
            city={Macau},
            postcode={999078},
            country={China}}

\affiliation[Sdu]{organization={Key Laboratory of Industrial Intelligent Systems},
            city={Jinan},
            postcode={250061},
            country={China}}
            
\author[Macau]{Wenmin Wang}
\ead{wmwang@must.edu.mo}

\author[sdu,SDu,Sdu]{Jinglin Zhang\corref{cor1}}
\ead{jinglin.zhang@sdu.edu.cn}

\begin{abstract}
Infrared and visible image fusion aims to generate a composite image that retains significant target information and preserves detailed textures, integrating two heterogeneous modalities. Previous image fusion methods typically adopt a single-module stacking approach to extract features from the two modalities. However, these approaches may result in incomplete learning of their distinct characteristics, thereby limiting the fusion effectiveness and constraining robustness in real-world heterogeneous data scenarios. To address these challenges, we propose FMRFusion, a frequency-aware multi-view representation learning network for Heterogeneous Image Fusion. A Multi-Scale Structural Perception Module is introduced to effectively capture discriminative structures, extracting fine-grained local structures and essential contextual information. A bilinear frequency decomposition mechanism is employed to separate features into high-frequency and low-frequency components, enabling joint modeling of local details and global representations across different frequency domains. Moreover, a Cross-View Complementary Interaction is incorporated to explicitly model and fuse the complementary characteristics between reflected light information and radiative intensity responses, facilitating effective cross-view interaction. We further improve the performance of the fused results by flow matching, which progressively refines the fused features by learning the transformation from coarse data to high-quality representations. Extensive experiments conducted on multiple benchmark datasets demonstrate that FMRFusion achieves superior and consistent performance across a range of fusion tasks, especially in nighttime scenarios.

\end{abstract}

\begin{keyword}
heterogeneous Image Fusion  \sep Frequency decomposition \sep  Cross-view complementary interaction \sep Flow matching
\end{keyword}

\end{frontmatter}


\section{Introduction}


Infrared and visible image fusion (IVIF) aims to integrate complementary information from heterogeneous views to enhance scene representation under low-quality imaging conditions \cite{xu2024imqfusion, liu2024infrared, zhao2026sfdfuse}.
The fused images are more compatible with human visual perception and benefit downstream vision tasks, such as object detection, tracking \cite{li2025mulfscap, dong2025you,qiang2026dwsfusion,Tang2026LoFTCLIP}, person re-identification, and semantic segmentation \cite{1,2,3,4,5,6}.
Visible images provide rich texture and color information in well-lit environments but suffer from severe degradation under low-light or adverse conditions.
In contrast, infrared sensors are robust to illumination variations by capturing thermal radiation, yet they typically lack high-frequency spatial details.
The fusion of these heterogeneous views enables more comprehensive and reliable representations, which are particularly important for vision tasks involving low-quality infrared observations.
Despite recent advances, existing methods still face challenges in jointly modeling global contextual information and preserving fine-grained details, especially when decoupling and learning complementary representations across modalities.
To address these challenges, we propose a frequency-aware multi-view learning framework for fusing infrared and visible images.

Prior to the advent of deep learning techniques \cite{jia2026multi}, several traditional algorithms for IVIF were proposed including Multiscale Transform, sparse representation, subspace approaches, and others.
These algorithms generally follow a pipeline that involves feature extraction based on predefined mathematical models, fusion via specific strategies, and subsequent reconstruction to obtain the final outputs. Despite achieving satisfactory performance, they often struggle to fully preserve the complementary information inherent in the source images.
In recent years, several approaches have been developed to tackle the Multiple-Modality Image Fusion challenge \cite{chen2026tofusion,ma2022swinfusion}.
Firstly, although CNNs excel at extracting local features, their small receptive fields make it challenging to model long-range dependencies, thereby limiting their ability to capture deep and global features of the source image \cite{zhang2021sdnet,zhang2020ifcnn}.
Secondly, methods specifically designed for infrared and visible image fusion must enhance multi-scale foundational information and detailed textures while extracting global features.
This capability cannot be achieved by the vast majority of Vision Transformer or CNN-based algorithms \cite{qu2022transmef,xu2022rfnet}.
Finally, most deep learning-based algorithms first connect the source images and feed them into a single-path network, without a separate feature extraction branch for each input image, leading to the loss of unique information \cite{long2021rxdnfuse}.
In summary, previous methods encounter three main limitations:
(1) \textbf{Incomplete feature decoupling}: Traditional MST-based and CNN-driven methods apply identical operators to both modalities, losing modality-specific features through shared encoding pipelines.
(2) \textbf{Local-global feature tradeoff}: CNN-based networks excel at local context but ignore long-range dependencies, while Vision Transformers achieve global modeling at prohibitive computational costs.
(3) \textbf{Suboptimal fusion mechanisms}: Single-path architectures concatenate inputs prematurely, causing information interference between modalities rather than structured feature interaction.

To address these limitations, we propose a frequency-aware multi-view representation learning network (FMRFusion) for low-quality infrared–visible image fusion. Our approach integrates the strengths of convolutional neural networks and Transformers, combining local context extraction and computational efficiency with global attention and long-range dependency modeling. Moreover, we introduce a feature decomposition module that separates image features into low-frequency and high-frequency components. Since infrared and visible images originate from the same scene, their low-frequency components capture shared information such as background structures, while high-frequency components encode modality-specific details, including textures in visible images and thermal patterns in infrared images. By explicitly modeling both shared and modality-specific features, FMRFusion enhances feature representation and improves fusion quality. In addition, we incorporate a Flow Matching-based refinement strategy to further enhance the fidelity and perceptual quality of the fused images.

The main contributions of this paper are summarized as follows:
\begin{itemize}
\item{\textbf{Dual-Branch Hybrid Architecture.} This work combines the efficiency of CNNs with the global modeling capabilities of Transformers to form independent branches for different modalities, thus avoiding feature interference during encoding.}

\item{\textbf{Frequency-Decoupled Fusion Mechanism.} Frequency-Decoupled Fusion Mechanism. We propose a novel feature decomposition module that explicitly separates and enhances shared structural information in low-frequency components and modality-specific details in high-frequency components..}

\item{\textbf{Flow Matching based Refinement} A flow matching paradigm is introduced to iteratively enhance the fused features by mapping coarse data to high-quality representations, yielding substantial improvements in fidelity and robustness.}

\end{itemize}
The rest of this paper is organized as follows. Section 2 briefly reviews related work on CNN-, AE-, GAN-, and Transformer-based fusion methods. Section 3 presents a comprehensive description of the proposed methodology. Section 4 reports experimental results and visual comparisons on infrared–visible image fusion. Section 5 presents experimental results and visual comparisons for medical image fusion. Section 6 presents experimental results and visual comparisons for multi-focus image fusion. Finally, Section 7 summarizes the effectiveness and future prospects of the proposed method.

\section{Related Work}
Based on the differences in deep neural network architectures and core learning paradigms, current mainstream infrared and visible image fusion methods can be systematically categorized into four key paradigms: fusion frameworks based on Convolutional Neural Networks, Autoencoders, Generative Adversarial Networks, and Transformers. This section provides a concise overview of representative methods under these paradigms.


\subsection{CNN-based methods}

In recent years, CNNs have dominated the field of image fusion due to their strong adaptability and effectiveness \cite{wang2025multi,li2026atdfusion,7,8,9,10,11,12}.
Typically, the image fusion process involves feature extraction, feature fusion, and image reconstruction.
To adaptively modulate the proportion of preserved gradient information and retain a more comprehensive texture structure, the SDNet \cite{zhang2021sdnet} framework has been proposed.To ensure gradient fidelity, this method sets the optimization objective based on texture richness, using an adaptive decision block to guide the fused image towards enhanced texture details.
FDAFusion \cite{cui2025fdafusion} introduces a cyclic feedback mechanism linking low-light enhancement, frequency-domain decomposition, and adaptive fusion for nighttime infrared-visible image fusion, and proposes FIDM and ACFM to capture cross-modal correlations, optimize feature weight allocation, and enhance the contrast, texture details, and visual fidelity of fused images in dim conditions.
DMNet \cite{pan2025dmnet} employs dense multi-scale feature extraction and dual attention guidance, adopting a two-stage training strategy of CNN encoder pre-training and diffusion model decoder fine-tuning to reinforce cross-modal complementary information mining and improve detail preservation and structural consistency of fused images.
U2Fusion \cite{xu2020u2fusion} integrates the VGG-16 network into its loss function for feature extraction, facilitating the fusion of multiple neural network architectures.

\subsection{AE-based methods}

The AE-based Models provide efficient feature representation and reconstruction capabilities for iamge fusion. Li et al. proposed DenseFuse \cite{li2018densefuse}, aiming to extract more valuable features from the source images. The fusion process culminates in the reconstruction of the fused image, which involves a manually designed fusion strategy and a decoding network.
Furthermore, they introduced NestFuse and RFN-Nest \cite{li2020nestfuse,li2021rfn} to preserve substantial information from the input data from a multi-scale perspective.
Specifically, the latter designed loss functions for detail preservation and feature enhancement to integrate more useful information from the source images.
DAE-Fuse \cite{park2025nerfies} proposes a two-phase discriminative autoencoder framework that mitigates modality bias and blurring artifacts while generating sharp and natural fused images. Furthermore, it extends image fusion from static images to temporally consistent video fusion.
To address the limitation of non-interpretable deep feature maps, CSF \cite{xu2021classification} adopts an unsupervised deep learning approach to achieve interpretable importance assessment of feature maps.

\subsection{GAN-based methods}
Generative Adversarial Networks \cite{zhang2025cpigan,ma2020ganmcc} are well-suited for unsupervised image fusion due to their strong ability to estimate probability distributions.
Ma et al. first engaged in an adversarial game between a generator and a discriminator, termed FusionGAN.
In their approach, the generator is continuously optimized through adversarial learning to obtain a fusion result that retains useful information from the source images, eliminating the need for manually designed fusion rules.
However, a single discriminator cannot account for the distribution of data across multiple domains.
To address this issue, DDcGAN, a Dual-Discriminator Conditional Generative Adversarial Network \cite{ma2020ddcgan}, was designed to simultaneously preserve the most important feature information in both infrared and visible images.
Building on DDcGAN, AttentionFGAN \cite{li2020attentionfgan} employs a multi-scale attention mechanism to preserve the foreground target information of infrared images and the rich background detail features of visible images.
R2GAN \cite{zhang2025cpigan} employs a recursive residual GAN architecture to progressively refine fused representations, enhancing salient targets and structural details through adversarial learning.
However, networks with two discriminators are difficult to converge.
In TarDAL \cite{liu2022target}, a bilevel optimization approach is introduced to the Target-aware Dual Adversarial Learning network, aimed at achieving long-term dependencies within source images.
Additionally, these networks often exhibit instability, making training challenging.

\subsection{Transformer-based methods}

In recent years, transformer-based approaches have been widely adopted for infrared and visible image fusion \cite{ding2025dskfuse, zhang2026sfdfuse,13,14,15,16,17,18}, benefiting from their ability to capture global context and long-range dependencies \cite{yan2026atfusion}.
SwinFusion proposes a generic image fusion framework based on cross-domain distance learning and the Swin Transformer \cite{ma2022swinfusion}.
SwinFuse constructs a comprehensive attention feature encoding backbone to model long-range dependencies \cite{wang2022swinfuse}.
DATFuse uses a dual-attention converter to extract important features while retaining global complementary information \cite{tang2023datfuse}.
TDFusion \cite{bai2025task} incorporates a learnable fusion loss guided by downstream task objectives to adaptively steer the fusion process, enabling the generation of task-oriented fused representations.
However, an autoencoder consisting solely of a Transformer cannot effectively extract local information.
For this reason, Zhao et al. \cite{zhao2021dndt} proposed a sequential DenseNet and Dual Transformer architecture to extract both local and global information, with the Dual Transformer enhancing the global information in the features before the fusion layer.
Subsequently, Li et al. \cite{li2022cgtf} proposed a convolutionally-guided converter that first learns local features using a convolutional feature extractor and then uses them to guide a converter-based feature extractor in capturing remote interdependencies of features.

\section{Methodology}

\subsection{Detail of FMRFusion}

As shown in \textbf{Fig.~\ref{fig:models}},  The encoder extracts and decomposes features from infrared and visible images, while the fusion layer integrates complementary information across modalities. A bilinear frequency decomposition mechanism separates features into high-frequency and low-frequency components to capture both local details and global structures. The decoder reconstructs the fused output, and the flow-matching module progressively refines the fused images to enhance fidelity and perceptual quality.

{
\vspace{12pt}  
\captionsetup[figure]{labelfont={bf},labelformat={default},labelsep=period,name={Fig.}}
\centering
\begin{minipage}{\linewidth}
  \centering
  \includegraphics[width=0.95\linewidth]{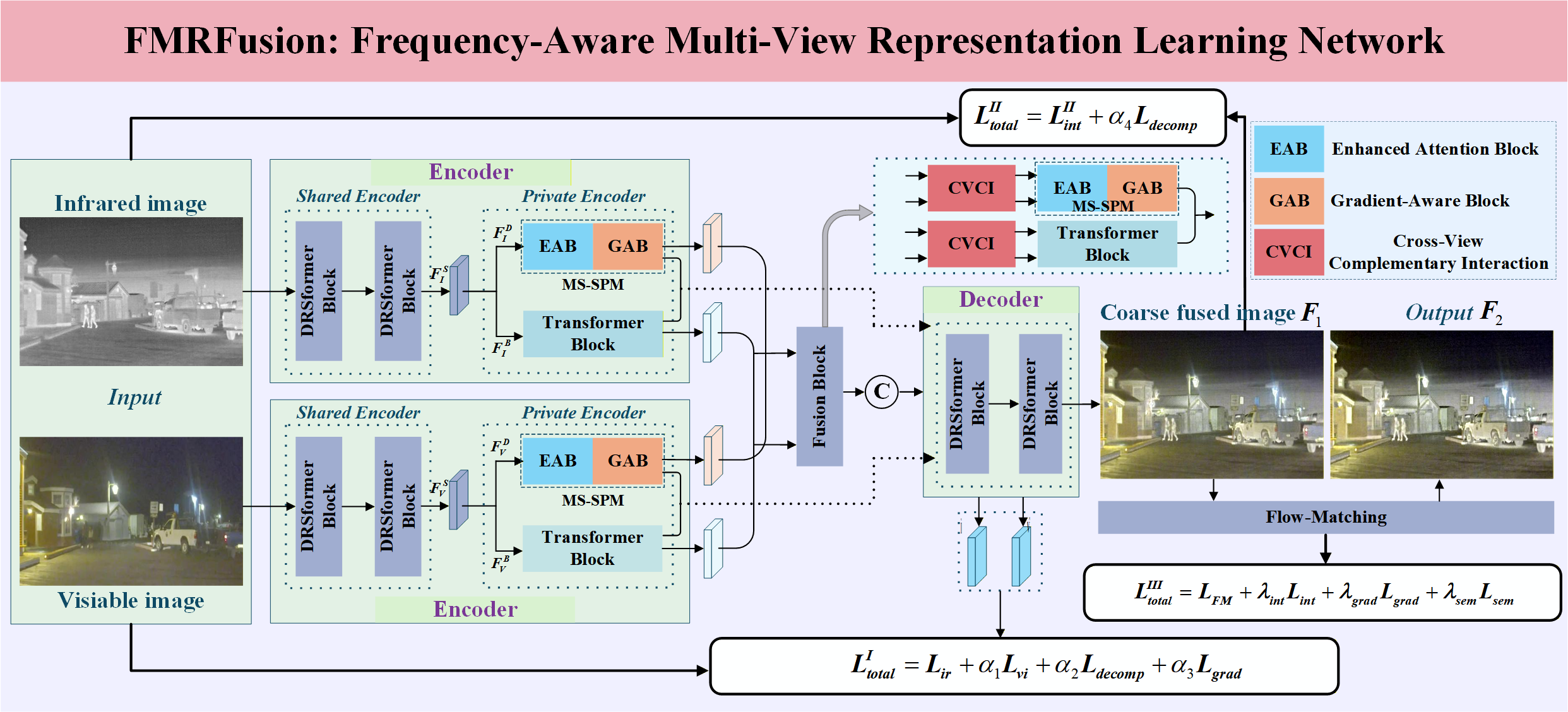}
  \captionof{figure}{Illustration of FMRFusion. In the feature decomposition stage, the infrared and visible images pairs $I$, $V$ are fed into the shallow DRSformer encoder SDE to obtain the features $F_I^S$,$F_V^S$, respectively. The feature $F_I^S$ is inputted into the Multi-Scale Structural Perception Module (MS-SPM) and Transformer Block (TRB) for feature decomposition, with the same process applied to $F_V^S$. It is worth mentioning that in MS-SPM, we use larger convolutional kernels to capture richer detail information within each modality. During the feature reconstruction stage, we reconstruct the infrared and visible images to enable the model to better capture modality-specific features. Eventually, the fused features are fed to the decoder to obtain the coarse fused image $F_1$. A flow-matching is used to enhance the coarse fused image to get the final fused result $F_2$.}
  \label{fig:models}
\end{minipage}
}

\subsubsection{Encoder}
The encoder consists of three components: the SDE, TRB, and MS-SPM.

Given a pair of infrared image $I$ and visible image $V$, the encoder extracts the shallow feature $F_I^S$,$ F_V^S$ from $I$, $V$. that is,
\begin{equation}
F_I^S = SDE(I), \quad F_V^S = SDE(V).
\end{equation}

Motivated by \cite{chen2023learning,jie2024tsjnet}, the DRSformer block uses a learnable parameter K to dynamically control the magnitude of sparsity, thus avoiding the inclusion of irrelevant information during the feature interaction process \cite{wang2022kvt}.
This enables the model to adaptively preserve the most useful self-attention values for improved feature aggregation.
Therefore, we incorporate DRSformer into the encoder to enhance its ability to facilitate complex cross-modal interactions between infrared and visible images.

The TRB is used to extract low-frequency information from shallow features.
Using Transformers with spatial self-attention better handles long-range dependencies, which is especially important for processing sequential data.
However, in pursuing better performance, we must also consider the balance between computational efficiency and the number of parameters.
For this purpose, we adopt the LT block as the fundamental unit of the TRB.
The LT block combines local and global attention mechanisms to enhance model performance by capturing local relationships within sequences and modeling long-range dependencies across sequences, which is formulated as:
\begin{equation}
F_I^B = T(F_I^S), \quad F_V^B = T(F_V^S).
\end{equation}
$F_I^B$ and $F_V^B$ are the base feature of $I$ and $V$, respectively. $T$(·) represents the TRB.

Due to significant differences in fine-grained features among different modal images, we propose a MS-SPM with a dual-branch structure to efficiently extract important local information.
The MS-SPM comprises two components: the Enhanced Attention Block (EAB) and the Gradient-Aware Block (GAB).

The specific design of the GAB is illustrated in \textbf{Fig.~\ref{fig:models}}.
The gradient-aware module is a variant of ResBlock, where the main stream utilizes dense connections while the residual stream operates with gradients.
In the main stream, two $3 \times 3$ convolutional layers and one $1 \times 1$ convolutional layer are deployed to fully utilize the features extracted by various convolutional layers.
We use the leaky rectified linear unit as the activation function. In the residual stream, a new gradient filter is introduced to compute the gradient magnitude of the features.
The gradient filter first employs a $3 \times 3$ convolutional layer followed by a Sobel operator in the horizontal direction to compute the magnitude of horizontal gradients.
Similarly, we can compute the magnitude of vertical gradients using another $3 \times 3$ convolutional layer followed by a vertical Sobel operator.
The gradient information is passed through a $1 \times 1$ convolutional layer to normalize the channel dimensions.
Then, the outputs of the main dense stream and the residual gradient stream are combined using element-wise addition to integrate deep features with fine-grained detail features.

The EAB aims to capture richer detail information in the image data, which is crucial for a comprehensive understanding of the input features.
Specifically, the EAB decomposes the traditional $21 \times 21$ large convolutional kernel into smaller $5 \times 5$ or $7 \times 7$ depthwise convolutional kernels, effectively reducing the parameter count and computational cost.
By utilizing larger convolutional kernels, the EAB can cover a broader image area, thereby capturing richer contextual information.
The incorporation of this global context enables the model to better understand local features and more accurately extract hidden important information from the image.
The expressions are as follows:

\begin{equation}
F_I^D = D(F_I^S), \quad F_V^D = D(F_V^S),
\end{equation}
where $F_I^D$ and $F_V^D$ are the detail feature of $I$ and $V$, respectively.$D$(·) represents the MS-SPM.

{
\vspace{12pt}  
\captionsetup[figure]{labelfont={bf},labelformat={default},labelsep=period,name={Fig.}}
\centering
\begin{minipage}{\linewidth}
  \centering
  \includegraphics[width=0.95\linewidth]{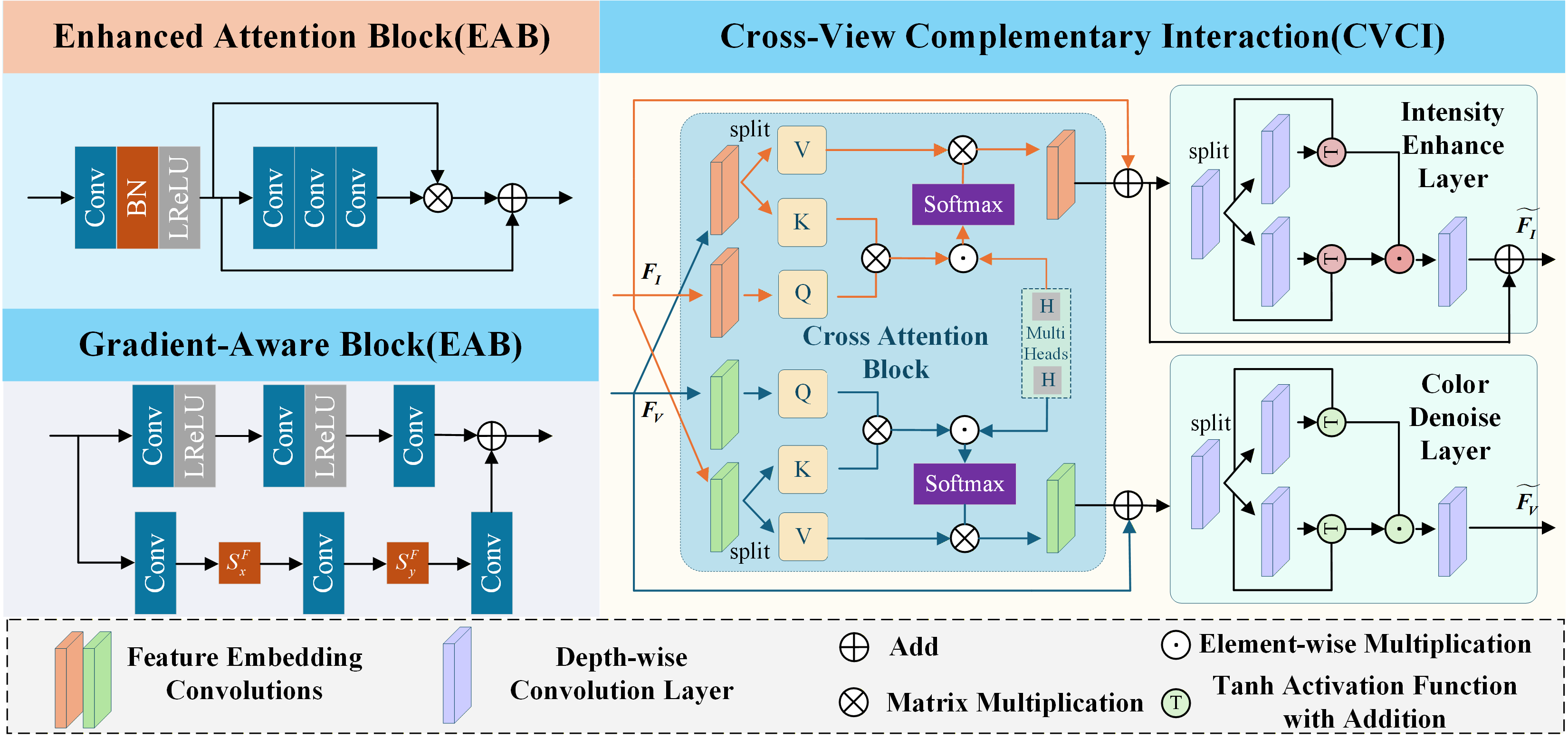}
  \captionof{figure}{Illustration of core modules in FMRFusion. The proposed framework consists of the EAB, GAB, and CVCI. The EAB enhances feature representation, the GAB strengthens structural details via gradient information, and the CVCI enables cross-modal interaction through cross-attention. 5Additionally, intensity enhancement and color denoising layers are employed to improve the quality of fused features.}
  \label{fig:EAB}
\end{minipage}
}


\subsubsection{Fusion Layer}

After feature decomposition, the detailed features extracted from the infrared image $I$ and the visible image $V$ are first fed into the MS-SPM for summation, while the global features corresponding to $I$ and $V$ are input into the TRB to model long-range dependencies. This process yields the pre-fused local and global features of the infrared and visible images, denoted as $F_D$ and $F_B$, respectively. The fusion process can be formulated as:
\begin{equation}
F_D = D(F_I^D + F_V^D), \quad
F_B = T(F_I^B + F_V^B),
\end{equation}
where $F_D$ and $F_B$ represent the fused local-detail features and global-salient features, respectively. Here, $D(\cdot)$ and $T(\cdot)$ denote the MS-SPM and TRB modules.

To further enhance the interaction between the structural information of infrared images and the brightness–color information of visible images, we introduce a Lighten Cross-Attention (LCA) module into the fusion stage. The LCA is designed to explicitly learn the complementary relationship between the I-branch, which mainly encodes infrared thermal structures and illumination cues, and the V-branch, which captures chromatic and appearance details from the visible image.

As illustrated in Fig.~\ref{fig:EAB}, the LCA module consists of a Cross-Attention Block (CAB), an Intensity Enhance Layer (IEL) for the I-branch, and a Color Denoise Layer (CDL) for the V-branch. The CAB adopts a symmetric dual-branch structure to enable cross-modal mutual guidance between infrared and visible features. To enforce effective cross-modal interaction, one branch is used to generate the query, while the features from the opposite modality are employed as the key and value, ensuring that each branch is refined exclusively using complementary information from the other modality.

Since the I-branch and V-branch share a similar architecture, we only describe  the I-branch for clarity. Let
\begin{equation}
F_I \in \mathbb{R}^{\hat{H} \times \hat{W} \times \hat{C}}
\end{equation}
denote the input intensity features derived from the infrared image. The CAB first computes the query as:
\begin{equation}
Q = W^{(Q)} F_I,
\end{equation}
while the key and value are obtained from the visible-guided features:
\begin{equation}
K = W^{(K)} F_V, \quad
V = W^{(V)} F_V,
\end{equation}
where $W^{(Q)}$, $W^{(K)}$, and $W^{(V)}$ denote feature embedding convolution layers. The output of the CAB is computed as:
\begin{equation}
\hat{F}_I = W\left( V \otimes \mathrm{Softmax}\left(Q \otimes K / \alpha_H \right) + F_I \right),
\end{equation}
where $\alpha_H$ is the multi-head scaling factor, $\otimes$ denotes matrix multiplication, and $W(\cdot)$ represents the feature embedding convolutions composed of a $1 \times 1$ depth-wise convolution followed by a $3 \times 3$ group convolution.

Following Retinex theory, the IEL further decomposes $\tilde{F}_I$ the intensity features into illumination and reflectance components:
\begin{equation}
F_I = W^{(I)} \hat{F}_I, \quad
F_R = W^{(R)} \hat{F}_I.
\end{equation}
The IEL is defined as:
\begin{equation}
\tilde{F}_I = W_s \left( \left(\tanh(W_s F_I) + F_I \right)
\odot
\left(\tanh(W_s F_R) + F_R \right) \right) + \hat{F}_I,
\end{equation}
where $\odot$ denotes element-wise multiplication and $W_s$ represents depth-wise convolution layers. Finally, a residual connection is applied to the IEL output to stabilize training and preserve infrared structural information. The CDL is symmetrically applied to the V-branch to suppress chromatic noise and maintain color consistency.

\subsubsection{Flow-Matching-Based Fusion Enhancement}
To enhance the fused image \cite{zhang2025sgdfuse}, a semantic-guided Flow Matching module is introduced. The preliminary fused image $F_1$ and the semantic masks $M_{ir}$ and $M_{vis}$ are concatenated as the semantic condition:
\begin{equation}
c=\mathrm{Concat}(F_1,M_{ir},M_{vis}).
\label{eq:fm_condition}
\end{equation}
where $F_1$ denotes the preliminary fused image obtained from Stage II. $M_{ir}$ and $M_{vis}$ represent the semantic masks extracted from the infrared and visible images, respectively. $c$ is the conditional guidance used to provide structural and semantic priors for the Flow Matching process.

During training, a Gaussian noise sample is drawn as:
\begin{equation}
z\sim\mathcal{N}(0,I).
\label{eq:fm_noise}
\end{equation}
where $z$ denotes the initial Gaussian prior, and $\mathcal{N}(0,I)$ represents a standard normal distribution with zero mean and identity covariance.

The intermediate state along the probability path is constructed as:
\begin{equation}
x_t=(1-t)z+tI_0,\quad t\in[0,1].
\label{eq:fm_path}
\end{equation}
where $x_t$ denotes the intermediate state at time step $t$. $I_0=\mathrm{Concat}(F_1,M_{ir},M_{vis})$ represents the semantically guided fusion representation, and $t$ controls the interpolation from the Gaussian prior to the fusion representation.

The target velocity field is defined as:
\begin{equation}
u_t=I_0-z.
\label{eq:fm_velocity}
\end{equation}
where $u_t$ denotes the target transport direction from the Gaussian prior $z$ to the semantically guided representation $I_0$.

A conditional U-Net is used to predict the velocity field:
\begin{equation}
v_\theta=f_\theta(x_t,t,c).
\label{eq:fm_predict}
\end{equation}
where $f_\theta(\cdot)$ denotes the conditional U-Net parameterized by $\theta$, and $v_\theta$ is the predicted velocity field conditioned on $x_t$, $t$, and $c$.

The Flow Matching loss is formulated as:
\begin{equation}
\mathcal{L}_{FM}
=
\mathbb{E}_{t,z,I_0}
\left[
\left\|
v_\theta(x_t,t,c)-u_t
\right\|_2^2
\right].
\label{eq:fm_loss}
\end{equation}
where $\mathcal{L}_{FM}$ constrains the predicted velocity field to match the target velocity field. $\mathbb{E}[\cdot]$ denotes the expectation over $t$, $z$, and $I_0$, and $\|\cdot\|_2$ represents the Euclidean norm.

During inference, the generation process starts from:
\begin{equation}
x_0\sim\mathcal{N}(0,I).
\label{eq:fm_infer_noise}
\end{equation}
where $x_0$ denotes the initial state sampled from the Gaussian prior.

The learned velocity field is integrated as:
\begin{equation}
\frac{dx}{dt}=v_\theta(x_t,t,c),
\label{eq:fm_ode}
\end{equation}
and can be numerically solved by:
\begin{equation}
x_{t+\Delta t}=x_t+\Delta t\,v_\theta(x_t,t,c).
\label{eq:fm_euler}
\end{equation}
where $\Delta t$ denotes the integration step size. The ODE progressively transports the initial noise toward the semantic-guided fusion representation.

The final enhanced fused image is obtained as:
\begin{equation}
F_2=x_1.
\label{eq:fm_output}
\end{equation}
where $F_2$ denotes the final enhanced fusion result produced at the end of the Flow Matching process.

{
\vspace{12pt}  
\captionsetup[figure]{labelfont={bf},labelformat={default},labelsep=period,name={Fig.}}
\centering
\begin{minipage}{\linewidth}
  \centering
  \includegraphics[width=0.95\linewidth]{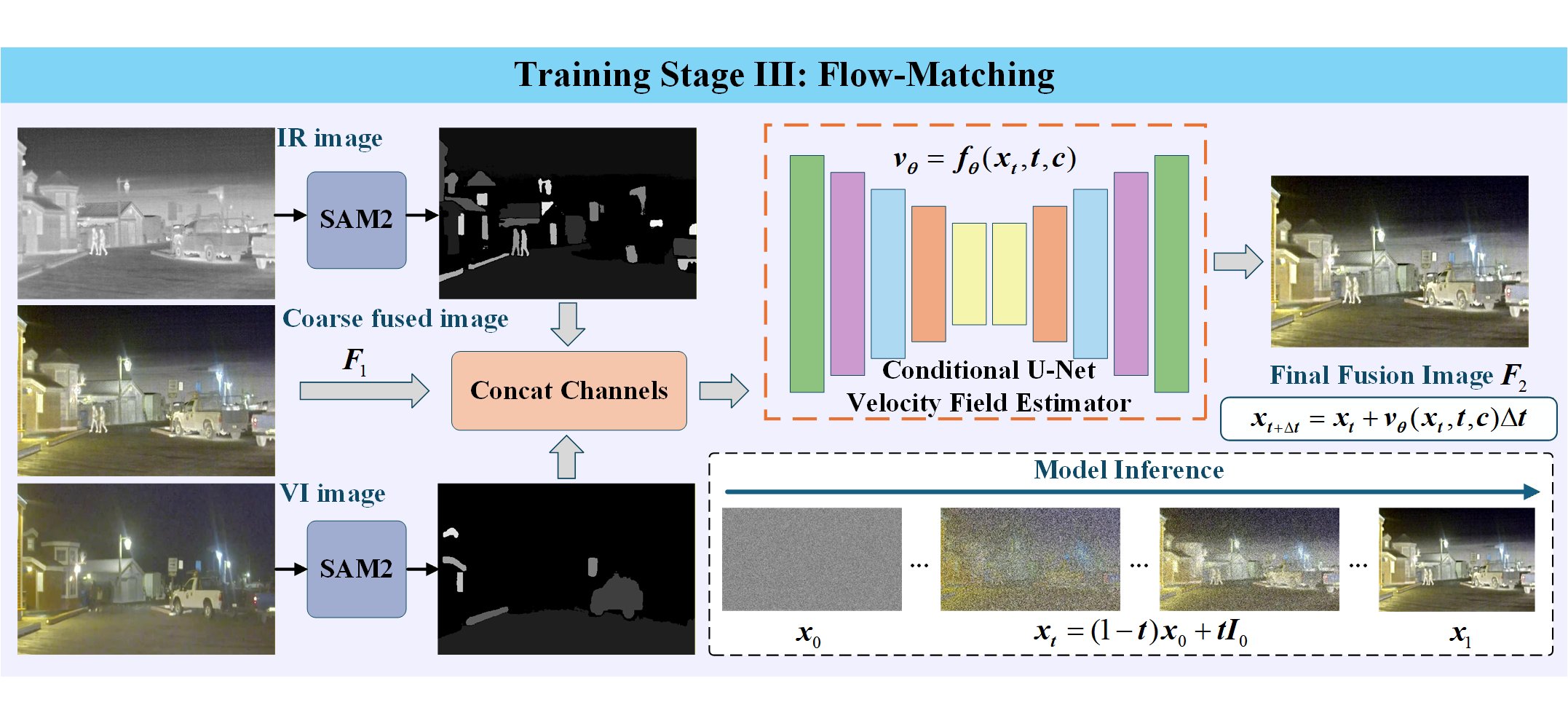}
  \captionof{figure}{Flow-Matching-Based Fusion Framework for Infrared–Visible Image Enhancement.Illustration of the Flow-Matching-based enhancement module applied to infrared–visible image fusion. The module progressively refines coarse Stage-II fusion results under semantic guidance using a conditional U-Net to estimate velocity fields. Key components include semantic masks generated from input images, linear interpolation to obtain intermediate states, and iterative Euler integration to produce the final enhanced fused image, improving saliency, texture preservation, contrast, and cross-modal alignment.}
  \label{fig:flow-matching}
\end{minipage}
}

\subsection{Loss Function}

We adopt a three-stage training strategy for FMRFusion, including feature decomposition, fusion learning, and flow-based enhancement.

\textbf{Stage I (Feature Decomposition and Reconstruction).}

Given infrared-visible image pairs $\{I, V\}$, shallow features are extracted and decomposed into base and detail components. The network is optimized via reconstruction, decomposition, and gradient constraints:
\begin{equation}
\mathcal{L}_{total}^{I} = \mathcal{L}_{ir} + \alpha_1 \mathcal{L}_{vi} + \alpha_2 \mathcal{L}_{decomp} + \alpha_3 \mathcal{L}_{grad}.
\end{equation}

where $\mathcal{L}_{ir}$ and $\mathcal{L}_{vi}$ denote the reconstruction losses for infrared and visible images, respectively. $\mathcal{L}_{decomp}$ represents the feature decomposition constraint, while $\mathcal{L}_{grad}$ denotes the gradient consistency loss. $\alpha_1$, $\alpha_2$, and $\alpha_3$ are weighting coefficients used to balance different loss terms.

The reconstruction losses are formulated as:
\begin{equation}
\begin{aligned}
\mathcal{L}_{ir} &= \| I - \hat{I} \|_2 + \theta (1 - SSIM(I, \hat{I})), \\
\mathcal{L}_{vi} &= \| V - \hat{V} \|_2 + \theta (1 - SSIM(V, \hat{V})).
\end{aligned}
\end{equation}

where $\hat{I}$ and $\hat{V}$ are the reconstructed images generated by the network. $\|\cdot\|_2$ denotes the Euclidean norm. $SSIM(\cdot,\cdot)$ represents the structural similarity index used to preserve structural information, and $\theta$ controls the contribution of the SSIM term.

The decomposition loss is defined as:
\begin{equation}
\mathcal{L}_{decomp} = \frac{(CC(F_I^{D}, F_V^{D}))^2}{CC(F_I^{B}, F_V^{B}) + \epsilon},
\end{equation}

where $F_I^{B}$ and $F_V^{B}$ denote the base features extracted from infrared and visible images, respectively, while $F_I^{D}$ and $F_V^{D}$ represent the corresponding detail features. $CC(\cdot,\cdot)$ denotes the correlation coefficient between feature maps. $\epsilon$ is a small constant introduced to avoid division instability. This loss encourages high correlation between base features and low correlation between detail features.

The gradient consistency loss is formulated as:
\begin{equation}
\mathcal{L}_{grad} = \frac{1}{HW} \left\| |\nabla I_f| - \max(|\nabla I_{ir}|, |\nabla I_{vi}|) \right\|_1.
\end{equation}

where $I_f$ denotes the fused image, and $I_{ir}$ and $I_{vi}$ denote the infrared and visible source images used for fusion. $\nabla(\cdot)$ represents the gradient operator, $|\cdot|$ denotes the gradient magnitude, and $\|\cdot\|_1$ denotes the $\ell_1$ norm. $\max(\cdot)$ is performed element-wise to preserve the strongest edge responses from both modalities. $H$ and $W$ denote the height and width of the image, respectively.

\textbf{Stage II (Fusion Learning).}

The decomposed features are fused and decoded to generate the fused image $I_f$, with the objective:
\begin{equation}
\mathcal{L}_{total}^{II} = \mathcal{L}_{int}^{II} + \alpha_4 \mathcal{L}_{decomp},
\end{equation}

where $\mathcal{L}_{int}^{II}$ denotes the intensity preservation loss and $\alpha_4$ is the weighting parameter for the decomposition constraint.

The intensity loss is defined as:
\begin{equation}
\mathcal{L}_{int}^{II} = \frac{1}{HW} \left\| I_f - \max(I_{ir}, I_{vi}) \right\|_1.
\end{equation}

where the fused image $I_f$ is encouraged to preserve the most salient intensity information from both infrared and visible images through the element-wise maximum operation.

\textbf{Stage III (Flow-Matching Enhancement).}

Since no ground-truth fused image is available, FMRFusion uses self-supervised constraints to guide the enhanced fusion result $F_2$. The total loss is formulated as:
\begin{equation}
\mathcal{L}_{total}^{III} = \mathcal{L}_{FM} + \lambda_{int} \mathcal{L}_{int} + \lambda_{grad} \mathcal{L}_{grad} + \lambda_{sem} \mathcal{L}_{sem},
\label{eq:sgdfuse_total_loss}
\end{equation}

where the individual terms are defined as:

\begin{equation}
\mathcal{L}_{int} = \| F_2 \odot M_{ir} - I \odot M_{ir} \|_1,
\end{equation}
which preserves salient infrared intensity, with $I$ denoting the infrared input image and $M_{ir}$ the infrared semantic mask.

\begin{equation}
\mathcal{L}_{grad} = \| \nabla F_2 \odot M_{vi} - \nabla V \odot M_{vi} \|_1,
\end{equation}
which preserves visible texture, with $V$ denoting the visible input image, $M_{vi}$ the visible semantic mask, and $\nabla$ the gradient operator.

\begin{equation}
\mathcal{L}_{sem} = \| F_2 \odot (M_{ir} + M_{vi}) - F_1 \odot (M_{ir} + M_{vi}) \|_1,
\end{equation}
which enforces semantic consistency between the enhanced fused image $F_2$ and the preliminary fusion $F_1$ from Stage II.

Here, $\lambda_{int}$, $\lambda_{grad}$, and $\lambda_{sem}$ are weighting factors balancing the contributions of intensity preservation, gradient preservation, and semantic consistency. The $\odot$ denotes element-wise multiplication to apply the mask. This design ensures that FMRFusion enhances $F_1$ with semantic guidance while preserving key infrared and visible features.

\section{Experiments}

\subsection{Setup}
\subsubsection{datasets}
We conducted experiments on four different datasets: the MSRS, TNO, Roadscene and M3FD datasets.
The MSRS dataset is a multispectral dataset comprising 1,083 pairs of images in the training set and 361 pairs in the test set.
In this experiment, we trained our network using the MSRS training dataset.
The TNO dataset is a widely used collection for IVIF, consisting of 60 pairs of images depicting various military-related scenes.
The RoadScene dataset consists of 221 pairs of registered infrared and visible images, featuring rich scenes of roads, vehicles, and pedestrians.
The M3FD dataset is designed for dual-spectral image fusion and detection, providing aligned visible and infrared data, and contains 300 pairs of fused images across from independent scenes.
In this experiment, we used the MSRS dataset (361 pairs), the TNO dataset (25 pairs), the RoadScene dataset (50 pairs), and the M3FD dataset (60 pairs) as our test datasets.

\subsubsection{Metrics}
We used nine metrics to quantitatively evaluate the fusion results, including entropy (EN), standard deviation (SD), spatial frequency (SF), visual information fidelity (VIF), mutual information (MI), structural similarity index (SSIM), sum of the correlations of differences (SCD), $Q^{AB/F}$, and average gradient (AG). Higher values indicate better fusion performance.

\subsubsection{Training details}
Our experiments were conducted on a machine equipped with an NVIDIA GeForce RTX 5090 GPU.
The number of training epochs was set to 240, with 80 epochs for the first stage and 160 epochs for the second stage.
The batch size is set to 8.
We use the Adam optimizer with the initial learning rate set to $10^{-4}$, decreasing by 0.5 every 20 epochs.
For the network hyperparameters, the number of DRSformer blocks in SDE is set to 4, with 8 attention heads and a dimensionality of 64.
The LT block in the TRB also has a dimensionality of 64 with 8 attention heads.
The decoder configuration is the same as that of the encoder.
To evaluate the performance of the proposed method, we conduct comparative experiments against several state-of-the-art multimodal image fusion networks, as summarized in the table. Specifically, the baseline methods include RCoNet (2022), LRRNet (2023) \cite{li2023lrrnet}, TarDAL (2022), BDLFusion (2023) \cite{liu2023bilevel}, CDDfuse (2023), Swinfuse (2022), and TDFusion (2025).

\captionsetup[figure]{labelfont={bf},labelformat={default},labelsep=period,name={Fig.}}
\captionsetup[table]{labelfont=bf,textfont=normalfont}
\definecolor{BestRed}{RGB}{255,200,200}
\definecolor{SecondBlue}{RGB}{200,220,255}
\newcommand{\best}[1]{\cellcolor{BestRed}{#1}}
\newcommand{\Second}[1]{\cellcolor{SecondBlue}{#1}}
\newcommand{\bestbox}[1]{\colorbox{BestRed}{\strut #1}}
\newcommand{\secondbox}[1]{\colorbox{SecondBlue}{\strut #1}}

We present qualitative comparisons in \textbf{Fig.~\ref{fig:MSRS} -~\ref{fig:M3FD}}.
In night scenes, infrared images highlight targets with active thermal radiation, while visible images provide rich color and texture only when there is sufficient light. Our method effectively integrates thermal radiation information from infrared images with detailed texture from visible images. Compared to most existing methods, our approach minimizes the loss of background information and effectively distinguishes between foreground targets and the background, resulting in more natural fusion results.

To comprehensively assess the quality of the generated images, we performed a quantitative comparison using image pairs from four datasets. The consolidated quantitative results for all metrics are presented in \textbf{Table~\ref{tab:quantitative_results}}.

First, on the \textbf{MSRS dataset}, our method exhibits excellent performance across four metrics. The highest EN indicates that our fused images contain more scene information, while the best SD demonstrates richer contrast and enhanced visual effects, attributed to our proposed Enhanced Attention Block. Furthermore, we achieved the best results in SF and AG, indicating that our outputs feature richer texture details and edge information. Although the VIF and Qabf metrics are not optimal, the differences compared to best models are minimal.

{
\vspace{12pt}  
\centering
\begin{minipage}{\linewidth}
  \centering
  \includegraphics[width=0.9\linewidth]{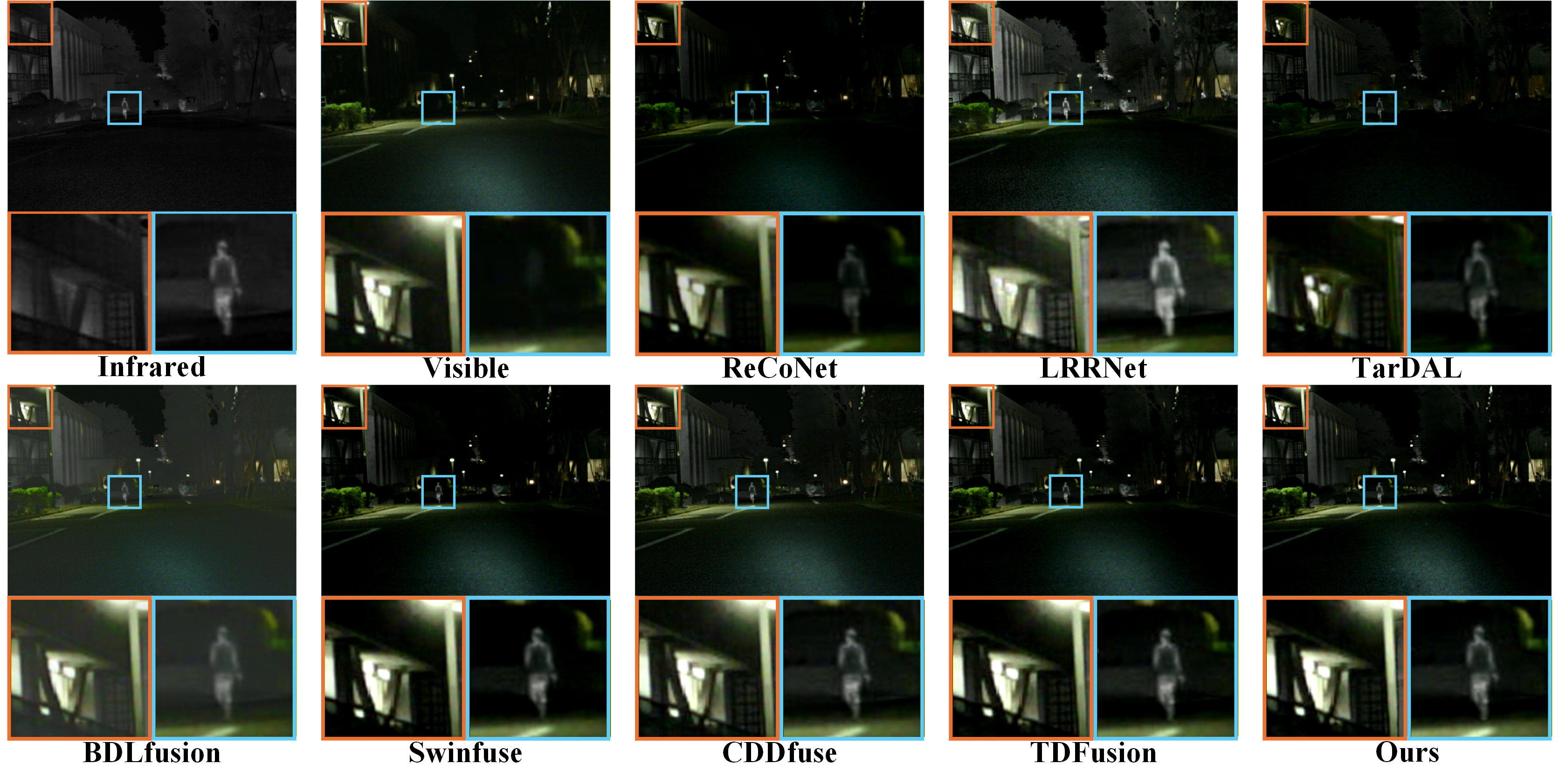}
  \captionof{figure}{Visual comparison for “01364N” in MSRS dataset.}
  \label{fig:MSRS}
\end{minipage}
}

\begin{figure}[H]
  \vspace{12pt}
  \centering
  \includegraphics[width=0.9\linewidth]{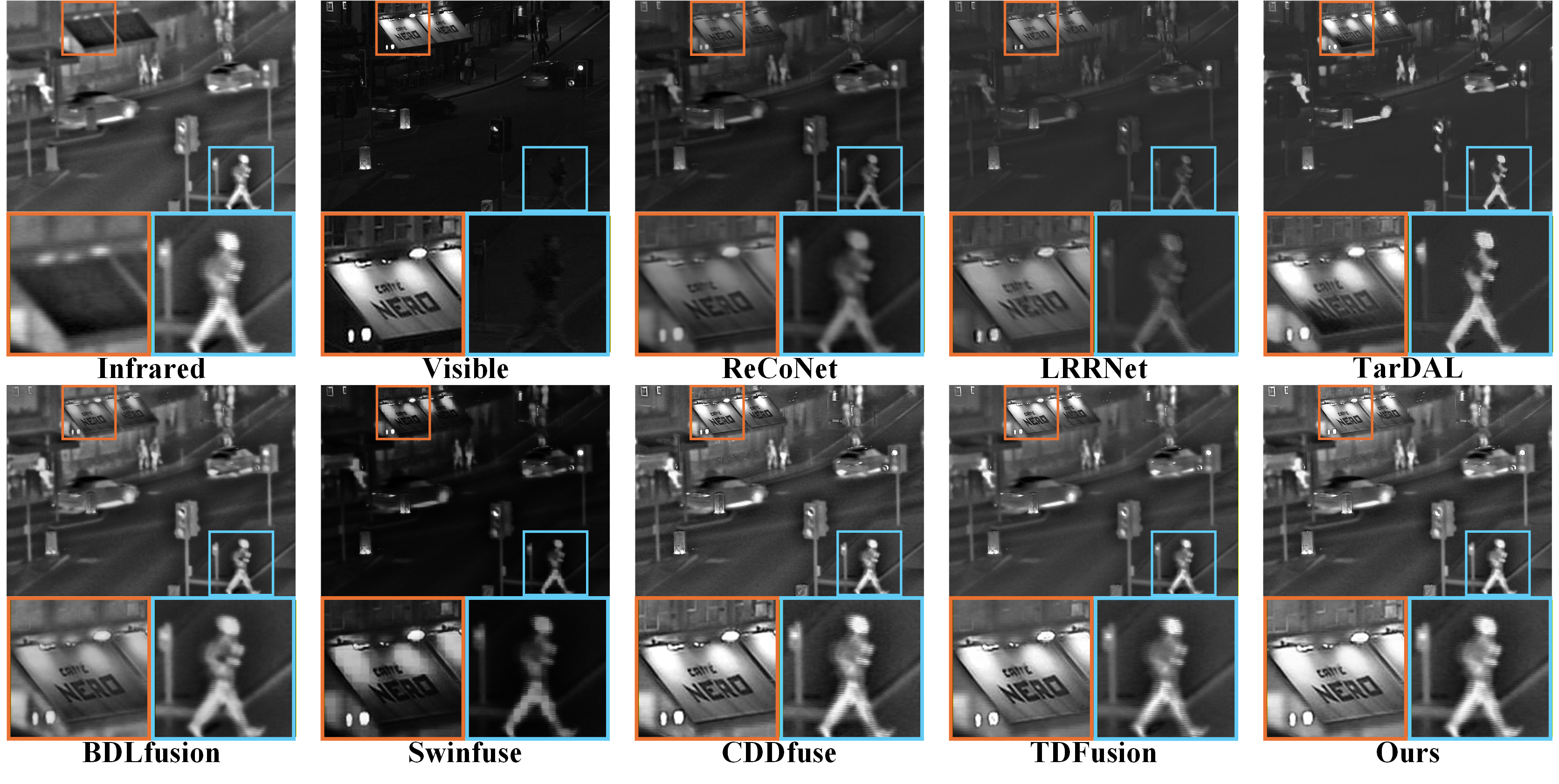}
  \caption{Visual comparison of ``2'' from the TNO dataset.}
  \label{fig:TNO}
\end{figure}
Second, regarding the \textbf{TNO dataset}, the proposed method delivers the best performance across the six evaluation metrics, demonstrating strong overall effectiveness. In particular, it obtains the highest scores on EN, SD, and SF, indicating that the fused images contain richer information content, higher contrast, and more abundant texture details. Moreover, our method also outperforms all competing approaches on VIF and $Q^{AB/F}$, reflecting superior visual fidelity and more effective preservation of source image information. In addition, the highest AG value further confirms the enhanced sharpness and structural clarity of the fusion results. These results indicate that the proposed method maintains robust and consistent performance on the TNO dataset, despite the relatively limited texture details in its visible images.

{
\vspace{12pt}  
\centering
\begin{minipage}{\linewidth}
  \centering
  \includegraphics[width=0.9\linewidth]{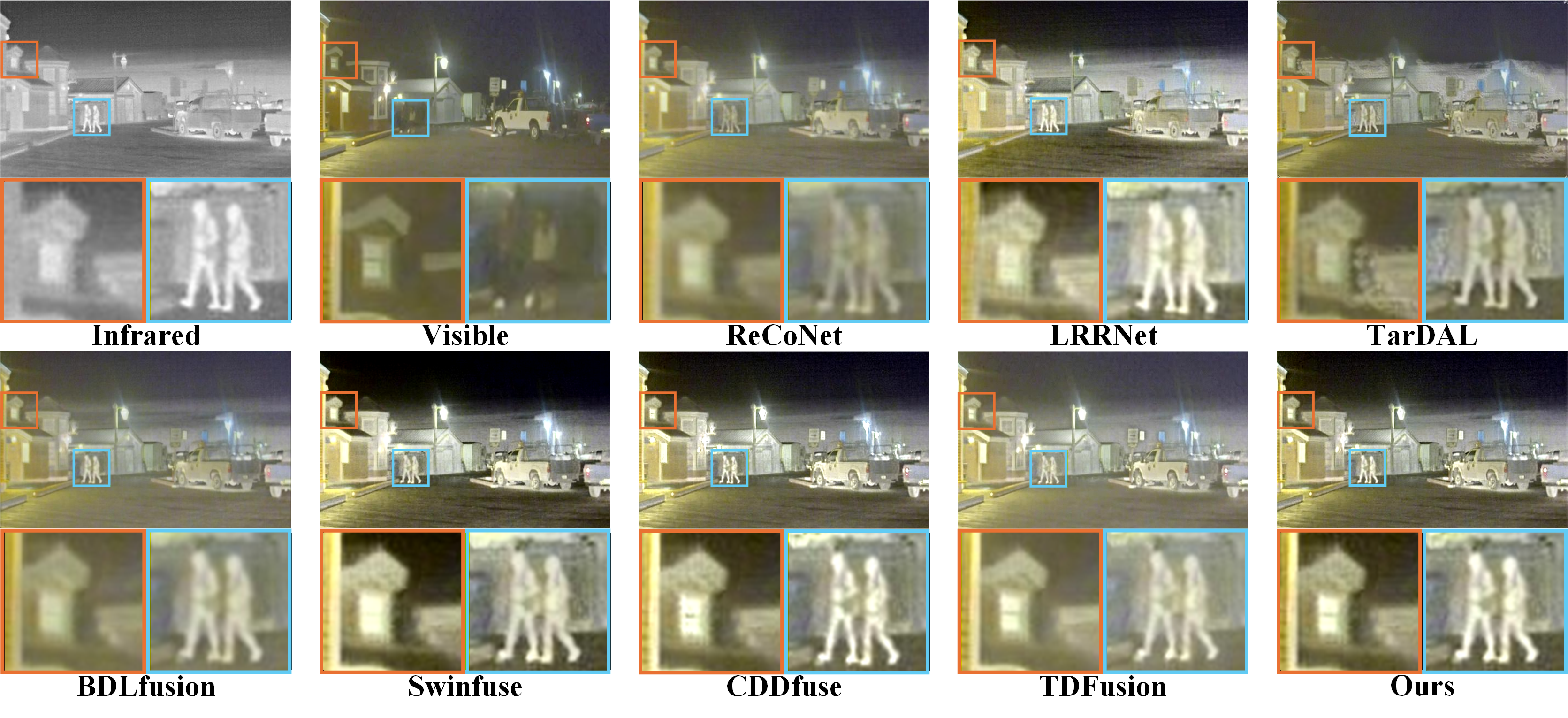}
  \captionof{figure}{Visual comparison for “FLIR 05857” in RoadScene dataset.}
  \label{fig:Roadscene}
\end{minipage}
}

{
\vspace{12pt}  
\centering
\begin{minipage}{\linewidth}
  \centering
  \includegraphics[width=0.9\linewidth]{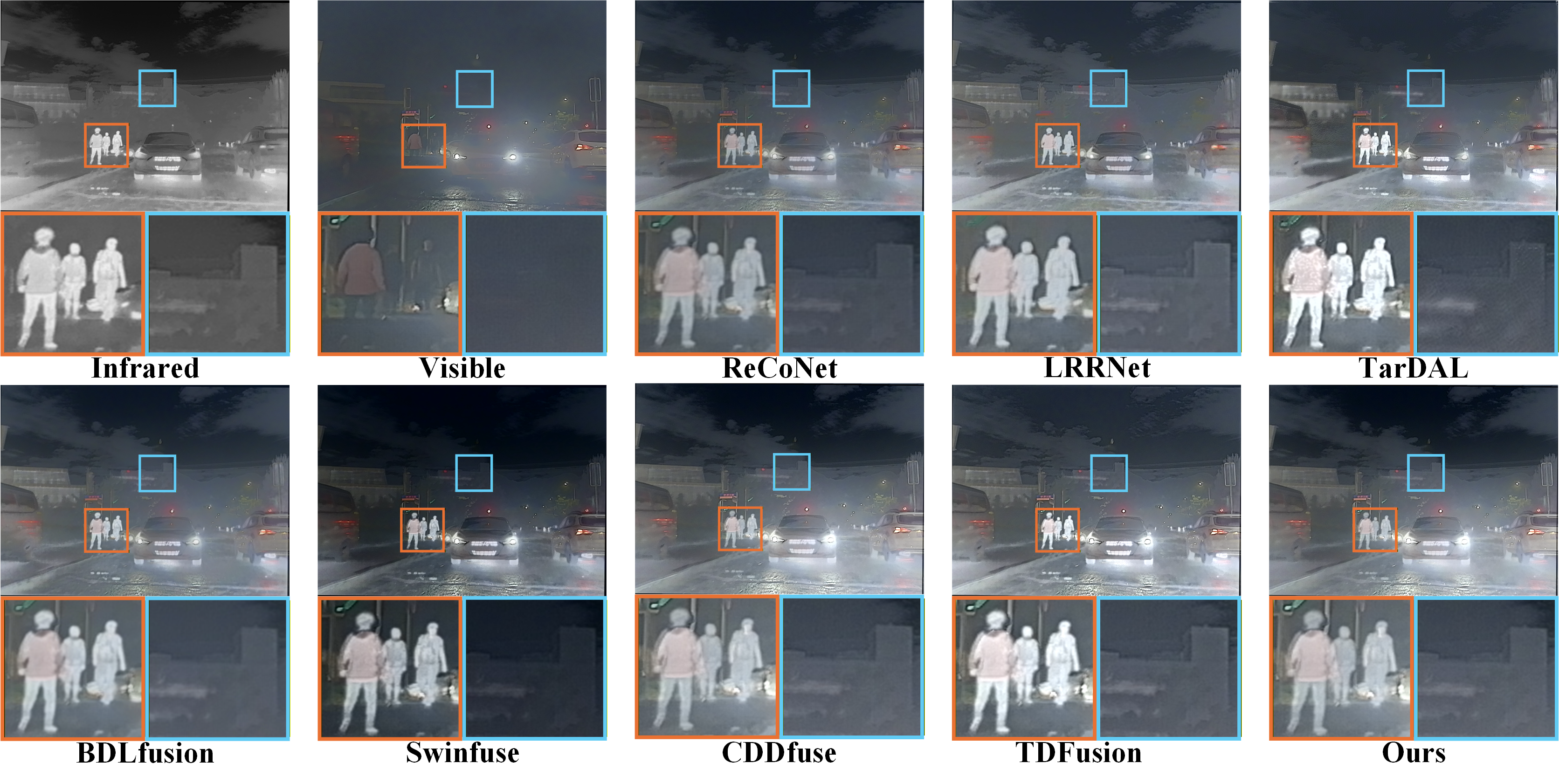}
  \captionof{figure}{Visual comparison for “00449” in M3FD dataset.}
  \label{fig:M3FD}
\end{minipage}
}

Third, for the \textbf{RoadScene dataset}, our method achieves the best performance across all evaluation metrics, demonstrating clear overall superiority. Specifically, it obtains the highest scores on SF and AG, indicating richer texture details and clearer structural information in the fused images. Our method also ranks first on SD, showing its strong ability to enhance image contrast and gray-level distribution. In addition, it achieves the highest EN value, reflecting that the fused results contain more abundant information. Moreover, our method outperforms all competing approaches on both VIF and $Q^{AB/F}$, demonstrating superior visual fidelity and more effective preservation of source image information. Overall, these results fully validate the robustness and effectiveness of the proposed method in complex road-scene scenarios.

Finally, for the \textbf{M3FD dataset}, our method achieves state-of-the-art performance across most evaluation metrics. Specifically, it obtains the best results in SD, SF, VIF, $Q^{AB/F}$, and AG, demonstrating strong capabilities in contrast enhancement, information preservation, and edge detail retention. Although CDDFuse achieves the highest EN value, our method attains the second-best EN while significantly outperforming all competing methods in SF and AG, which indicates superior texture representation and local activity preservation. These results confirm that our approach can effectively fuse infrared and visible images while maintaining rich structural and textural details in complex and diverse scenes.

\begin{table*}[h!]
\centering
\captionsetup{font=small}
\caption{Quantitative comparison of IVIF datasets.
The \bestbox{red} and \secondbox{blue} markers represent the best and second-best values.}
\label{tab:quantitative_results}
\footnotesize
\setlength{\tabcolsep}{4pt}
\renewcommand{\arraystretch}{1.15}

\begin{tabular}{cc}

\begin{tabular}{lcccccc}
\multicolumn{7}{c}{\textbf{MSRS Dataset}} \\
\hline
Method & EN$\uparrow$ & SD$\uparrow$ & SF$\uparrow$ & VIF$\uparrow$ & $Q^{AB/F}\uparrow$ & AG$\uparrow$ \\
\hline
ReCoNet   & 4.233 & 41.715 & 9.975  & 0.490 & 0.403 & 3.000 \\
LRRNet    & 6.191 & 31.757 & 8.473  & 0.541 & 0.454 & 2.650 \\
TarDAL    & 6.209 & 32.806 & 9.540  & 0.631 & 0.415 & 2.937 \\
BDLFusion & 6.093 & 33.629 & 7.796  & 0.726 & 0.497 & 2.599 \\
SwinFusion& 6.621 & 42.998 & 11.088 & 0.990 & 0.654 & 3.566 \\
CDDFuse   & 6.694 & 43.011 & \Second{11.632} & \best{1.066} & \best{0.714} & 3.735 \\
TDFusion  & \Second{6.740} & \Second{43.536} & 11.300 & 0.500 & 0.670 & \Second{3.845} \\
Ours      & \best{6.907} & \best{48.539} & \best{13.486} & \Second{1.056} & \Second{0.695} & \best{4.538} \\
\hline
\end{tabular}

&

\begin{tabular}{lcccccc}
\multicolumn{7}{c}{\textbf{TNO Dataset}} \\
\hline
Method & EN$\uparrow$ & SD$\uparrow$ & SF$\uparrow$ & VIF$\uparrow$ & $Q^{AB/F}\uparrow$ & AG$\uparrow$ \\
\hline
ReCoNet   & 6.677 & 40.457 & 7.957  & 0.530 & 0.372 & 3.367 \\
LRRNet    & 6.898 & 38.571 & 9.437  & 0.538 & 0.366 & 3.641 \\
TarDAL    & 6.723 & 39.113 & 11.378 & 0.543 & 0.404 & 4.090 \\
BDLFusion & 6.955 & 39.502 & 8.454  & 0.645 & 0.452 & 3.467 \\
SwinFusion& 6.899 & 41.172 & 11.371 & 0.757 & 0.538 & 4.388 \\
CDDFuse   & \Second{7.048} & \Second{44.984} & \Second{13.509} & \Second{0.809} & \Second{0.556} & \Second{4.947} \\
TDFusion  & 6.978 & 42.376 & 13.102 & 0.785 & 0.498 & 4.682 \\
Ours      & \best{7.164} & \best{45.988} & \best{15.663} & \best{0.898} & \best{0.575} & \best{5.832} \\
\hline
\end{tabular}

\\[8pt]

\begin{tabular}{lcccccc}
\multicolumn{7}{c}{\textbf{RoadScene Dataset}} \\
\hline
Method & EN$\uparrow$ & SD$\uparrow$ & SF$\uparrow$ & VIF$\uparrow$ & $Q^{AB/F}\uparrow$ & AG$\uparrow$ \\
\hline
ReCoNet   & 7.029 & 41.865 & 8.472  & 0.558 & 0.398 & 3.457 \\
LRRNet    & 6.990 & 39.889 & 11.123 & 0.494 & 0.347 & 4.032 \\
TarDAL    & 7.051 & 44.400 & 11.316 & 0.585 & 0.451 & 4.204 \\
BDLFusion & 6.981 & 36.448 & 8.764  & 0.584 & 0.424 & 3.553 \\
SwinFusion& \Second{7.238} & 45.000 & 11.553 & 0.677 & 0.502 & 4.292 \\
CDDFuse   & 7.125 & \Second{50.664} & \Second{16.390} & \Second{0.693} & \Second{0.542} & \Second{5.875} \\
TDFusion  & 6.956 & 42.581 & 12.584 & 0.599 & 0.421 & 4.806 \\
Ours      & \best{7.536} & \best{51.653} & \best{16.617} & \best{0.751} & \best{0.570} & \best{6.044} \\
\hline
\end{tabular}

&

\begin{tabular}{lcccccc}
\multicolumn{7}{c}{\textbf{M3FD Dataset}} \\
\hline
Method & EN$\uparrow$ & SD$\uparrow$ & SF$\uparrow$ & VIF$\uparrow$ & $Q^{AB/F}\uparrow$ & AG$\uparrow$ \\
\hline
ReCoNet   & 6.472 & 26.899 & 8.243  & 0.522 & 0.314 & 2.924 \\
LRRNet    & 6.893 & 39.870 & 11.123 & 0.629 & 0.243 & 4.132 \\
TarDAL    & 6.634 & 39.364 & 6.942  & 0.285 & 0.291 & 3.501 \\
BDLFusion & 7.110 & 35.998 & 6.969  & 0.425 & 0.334 & 3.935 \\
SwinFusion& 7.126 & 38.214 & \Second{17.055} & 0.633 & 0.550 & 5.784 \\
CDDFuse   & \best{7.325} & \Second{40.884} & 16.390 & \Second{0.693} & 0.542 & \Second{5.875} \\
TDFusion  & 6.990 & 40.633 & 14.490 & 0.610 & \Second{0.650} & 5.346 \\
Ours      & \Second{7.194} & \best{42.870} & \best{18.664} & \best{0.859} & \best{0.717} & \best{6.242} \\
\hline
\end{tabular}

\end{tabular}
\end{table*}

\subsection{Ablation studies}

As shown in \textbf{Table~\ref{tab:ablation_components_9m}}, we evaluate the necessity of EAB and GAB blocks in the MS-SPM.
We modified the MS-SPM section to use only GAB modules, meaning that detail features are extracted solely by GAB.
The results indicate that removing the EAB blocks from FMRFusion hinders the model's ability to fully utilize richer features for reconstructing clean visible images, resulting in decreased fusion performance compared to our complete model.
Similarly, in Experiment II, we extracted modality detail features solely using the EAB blocks.
The results demonstrate that this module enables FMRFusion to reconstruct finer detail features, thereby contributing more effective and subtle texture details to the fusion process, ultimately enhancing the quality of the fusion results.
Subsequently, in Experiment III, we replaced the DRSformer module with a transformer module composed of Restormer blocks with similar parameters \cite{zamir2022restormer}.
The results were slightly inferior overall compared to those obtained with the DRSformer module, demonstrating that the DRSformer enhances shallow feature extraction. In Experiment IV, the performance degradation across all metrics indicates that CVCI facilitates the integration of infrared thermal responses and visible structural details.

\begin{table*}[h!]
  \centering
  \captionsetup{font=small}
  \footnotesize
  \setlength{\tabcolsep}{4pt}
  \renewcommand{\arraystretch}{1.15}
  \caption{Ablation experiment results on the MSRS test set. Bold indicates the best value.}
  \label{tab:ablation_components_9m}
  \setlength{\tabcolsep}{6pt}
  \renewcommand{\arraystretch}{1.15}
  \begin{tabular}{ccccccccccc}
    \hline
    \multirow{2}{*}{}
    & \multirow{2}{*}{Configurations}
    & \multirow{2}{*}{EN$\uparrow$}
    & \multirow{2}{*}{SD$\uparrow$}
    & \multirow{2}{*}{SF$\uparrow$}
    & \multirow{2}{*}{MI$\uparrow$}
    & \multirow{2}{*}{SSIM$\uparrow$}
    & \multirow{2}{*}{SCD$\uparrow$}
    & \multirow{2}{*}{VIF$\uparrow$}
    & \multirow{2}{*}{$Q^{AB/F}\uparrow$}
    & \multirow{2}{*}{AG$\uparrow$} \\
    & & & & & & & & & & \\
    \hline
    I   & w/o EAB
        & 6.692 & 42.926 & 11.785
        & 2.361 & 0.646 & 1.421 & 0.958 & 0.648 & 3.887 \\
    II  & w/o GAB
        & 6.774 & 46.122 & 12.378
        & 2.514 & 0.668 & 1.507 & 1.001 & 0.659 & 4.070 \\
    III & DRSformer $\rightarrow$ Restormer
        & 6.760 & 46.642 & 12.540
        & 2.540 & 0.662 & 1.526 & 1.008 & 0.675 & 4.052 \\
    IV  & w/o CVCI
        & 6.785 & 48.129 & 13.191
        & 2.689 & 0.670 & 1.588 & 1.010 & 0.685 & 4.287 \\
    \hline
        & \textbf{Ours}
        & \textbf{6.907} & \textbf{48.539} & \textbf{13.486}
        & \textbf{2.752} & \textbf{0.685} & \textbf{1.623}
        & \textbf{1.056} & \textbf{0.695} & \textbf{4.538} \\
    \hline
  \end{tabular}
\end{table*}

\begin{table*}[h!]
  \centering
  \captionsetup{font=small}
  \footnotesize
  \setlength{\tabcolsep}{4pt}
  \renewcommand{\arraystretch}{1.15}
  \caption{The results of hyperparameter ablation experiments on the MSRS test set. 
  Bold indicates the best value.}
  \label{tab5}
  \setlength{\tabcolsep}{7pt} 
  \renewcommand{\arraystretch}{1.1}
  \begin{tabular}{ccccccccccc}
  \hline
   \multirow{2}{*}{} 
    & \multirow{2}{*}{Configurations}
    & \multirow{2}{*}{EN$\uparrow$}
    & \multirow{2}{*}{SD$\uparrow$}
    & \multirow{2}{*}{SF$\uparrow$}
    & \multirow{2}{*}{MI$\uparrow$}
    & \multirow{2}{*}{SSIM$\uparrow$}
    & \multirow{2}{*}{SCD$\uparrow$}
    & \multirow{2}{*}{VIF$\uparrow$}
    & \multirow{2}{*}{$Q^{AB/F}\uparrow$}
    & \multirow{2}{*}{AG$\uparrow$} \\
    & & & & & & & & & & \\
  \hline
    I   
    & [$\frac{1}{5}$ , $\frac{1}{2}$] 
    & 6.779 & 47.415 & 13.037 
    & 2.631 & 0.667 & 1.551 & 1.013 & 0.684
    & 4.198 \\

    II  
    & [$\frac{1}{6}$ , $\frac{1}{3}$] 
    & 6.765 & 46.497 & 12.829 
    & 2.590 & 0.646 & 1.541 & 0.938 & 0.646
    & 4.204 \\

    III 
    & [$\frac{2}{3}$ , $\frac{5}{6}$] 
    & 6.760 & 46.642 & 12.540 
    & 2.599 & 0.612 & 1.498 & 0.936 & 0.624
    & 4.052 \\
  \hline
        & \textbf{Ours} 
        & \textbf{6.907} 
        & \textbf{48.539} 
        & \textbf{13.486} 
        & \textbf{2.752}
        & \textbf{0.685}
        & \textbf{1.623}
        & \textbf{1.056}
        & \textbf{0.695}
        & \textbf{4.538} \\
  \hline
  \end{tabular}
\end{table*}

$\textbf{Effect of the number of K.}$ The adjustable parameter K in SDE is a controlled range of intervals that can dynamically learn the score that contributes the most. When K is too small, the lack of sufficient global information aggregation will undoubtedly lead to a sharp decline in performance. As shown in \textbf{Table~\ref{tab5}}, we test the model effect in three [$\Delta 1$, $\Delta 2$] ranges, and the results show that it works best in [$\frac{1}{2}$ , $\frac{4}{5}$].

\begin{table*}[h!]
  \centering
  \captionsetup{font=small}
  \footnotesize
  \setlength{\tabcolsep}{4pt}
  \renewcommand{\arraystretch}{1.15}  
  \caption{Results of Three-stage training strategy ablation experiments on the MSRS test set. Bold indicates the best performance.}
  \label{tab:ablation_9metrics}
  \setlength{\tabcolsep}{7pt}
  \renewcommand{\arraystretch}{1.15}
  \begin{tabular}{c c c c c c c c c c c}
    \hline
    \multirow{2}{*}{} 
    & \multirow{2}{*}{Configurations}
    & \multirow{2}{*}{EN$\uparrow$}
    & \multirow{2}{*}{SD$\uparrow$}
    & \multirow{2}{*}{SF$\uparrow$}
    & \multirow{2}{*}{MI$\uparrow$}
    & \multirow{2}{*}{SSIM$\uparrow$}
    & \multirow{2}{*}{SCD$\uparrow$}
    & \multirow{2}{*}{VIF$\uparrow$}
    & \multirow{2}{*}{$Q^{AB/F}\uparrow$}
    & \multirow{2}{*}{AG$\uparrow$} \\
    & & & & & & & & & & \\
    \hline
    I  & w/o first stage  
       & 6.752 & 45.637 & 12.615 
       & 2.479 & 0.674 & 1.482 & 0.982 & 0.661 & 4.108 \\
    II & w/o second stage 
       & 4.851 & 28.104 & 8.475  
       & 2.046 & 0.617 & 1.203 & 0.901 & 0.612 & 2.279 \\
    III  & w/o three stage  
       & 6.807 & 46.832 & 13.072
       & 2.533 & 0.680 & 1.579 & 0.995 & 0.680 & 4.327 \\    
    \hline
       & \textbf{Ours} 
       & \textbf{6.907} & \textbf{48.539} & \textbf{13.486} 
       & \textbf{2.752} & \textbf{0.685} & \textbf{1.623} 
       & \textbf{1.056} & \textbf{0.695} & \textbf{4.538} \\
    \hline
  \end{tabular}
\end{table*}

$\textbf{Three-stage training strategy.}$ To verify the necessity of the three-stage training approach, we ablate each stage, as shown in \textbf{Table~\ref{tab:ablation_9metrics}}. Experimental results indicate that using incomplete training configurations leads to unsatisfactory performance on all four evaluation metrics of the MSRS dataset. In contrast, the complete three-stage training strategy achieves significant performance improvements.

\section{Medical image fusion}
To validate the generalization of the FMRFusion, we directly tested the two datasets of medical images using the training weights from the MSRS dataset. We selected 286 pairs of medical images from the Harvard Medical website for image fusion experiments, including 21 pairs of MRI-CT images, 42 pairs of MRI-PET images, and 73 pairs of MRI-SPECT images as the test dataset. We adopt the same evaluation metrics as those used for infrared–visible image fusion to ensure consistency across experiments.


Qualitative and quantitative results are presented in \textbf{Fig.~\ref{fig:PET}} and \textbf{Table~\ref{tab:mri_ct_pet_results}}. FMRFusion successfully preserves histologic information in MRI and functional information in PET, both of which are crucial for diagnosis and disease monitoring. This capability is attributed to the design of the dual-branch feature decomposition network, which effectively integrates information from different modalities, thereby enhancing the overall information content of the images. Moreover, our Three-stage training strategy optimizes the model's generalization ability across diverse medical imaging scenarios, allowing it to maintain high performance even in complex cases.

{


\begin{table*}[h!]
\centering
\captionsetup{font=small}
\caption{Fusion quality evaluation on the Medical datasets.
The \bestbox{red} and \secondbox{blue} markers represent the best and second-best values.}
\label{tab:mri_ct_pet_results}

\footnotesize
\setlength{\tabcolsep}{4pt}
\renewcommand{\arraystretch}{1.15}

\begin{tabular}{cc}

\begin{tabular}{lcccccc}
\multicolumn{7}{c}{\textbf{MRI-CT Dataset}} \\
\hline
Method & EN$\uparrow$ & SD$\uparrow$ & SF$\uparrow$ & VIF$\uparrow$ & Qabf$\uparrow$ & AG$\uparrow$ \\
\hline
ReCoNet   & 4.359 & 62.106 & 18.687 & 0.397 & 0.399 & 5.968 \\
LRRNet    & 4.716 & 37.235 & 16.656 & 0.350 & 0.306 & 4.855 \\
TarDAL    & 5.075 & 60.938 & 21.949 & 0.465 & 0.381 & 5.007 \\
BDLFusion & 4.733 & 74.154 & 19.829 & 0.479 & 0.462 & 5.386 \\
SwinFusion& 5.088 & 79.010 & 26.484 & 0.542 & 0.587 & 7.685 \\
CDDFuse   & \Second{5.106} & \Second{82.148} & \Second{39.367} & 0.494 & 0.616 & \Second{9.720} \\
TDFusion & 5.101 & 76.589 & 35.734 & \best{0.566} & \best{0.634} & 8.733 \\
Ours      & \best{5.245} & \best{84.479} & \best{44.355} & \Second{0.554} & \Second{0.633} & \best{10.882} \\
\hline
\end{tabular}

&

\begin{tabular}{lcccccc}
\multicolumn{7}{c}{\textbf{MRI-PET Dataset}} \\
\hline
Method & EN$\uparrow$ & SD$\uparrow$ & SF$\uparrow$ & VIF$\uparrow$ & Qabf$\uparrow$ & AG$\uparrow$ \\
\hline
ReCoNet   & 3.158 & 59.131 & 13.611 & 0.472 & 0.295 & 4.170 \\
LRRNet    & 4.355 & 48.295 & 13.397 & 0.366 & 0.207 & 3.937 \\
TarDAL    & \best{4.697} & 58.857 & 18.778 & 0.578 & 0.481 & 5.273 \\
BDLFusion & 4.380 & 60.879 & 16.759 & 0.549 & 0.528 & 5.282 \\
SwinFusion& 4.542 & 74.906 & 22.434 & \best{0.703} & 0.649 & 6.838 \\
CDDFuse   & 4.500 & \Second{78.165} & \Second{28.272} & 0.656 & 0.678 & \Second{7.622} \\
TDFusion  & 4.386 & 75.845 & 27.168 & 0.589 & \best{0.686} & 7.263 \\
Ours      & \Second{4.561} & \best{80.158} & \best{30.829} & \Second{0.689} & \Second{0.682} & \best{8.394} \\
\hline
\end{tabular}

\end{tabular}
\end{table*}

\vspace{12pt}  
\captionsetup[figure]{labelfont={bf},labelformat={default},labelsep=period,name={Fig.}}
\centering
\begin{minipage}{\linewidth}
  \centering
  \includegraphics[width=0.9\linewidth]{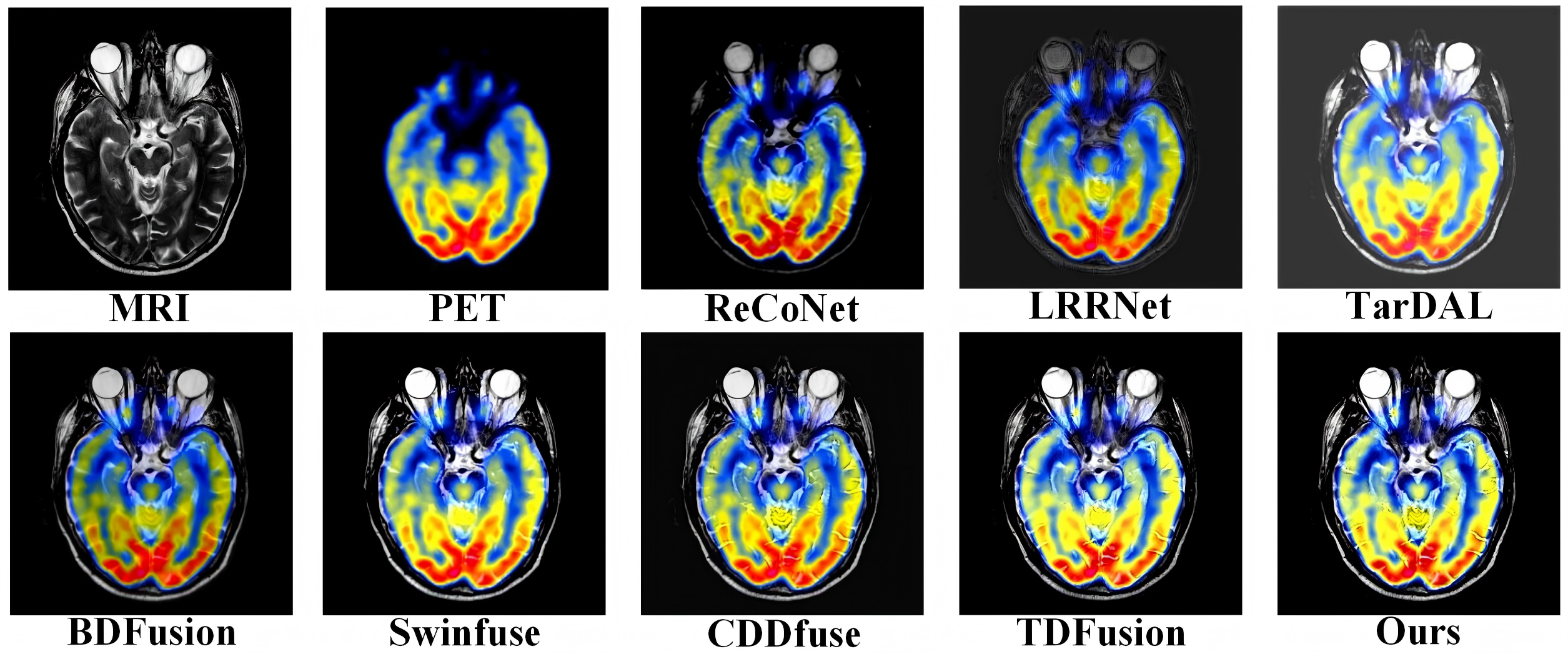}
  \captionof{figure}{Visual comparison for“MRI-PET-28”in MRI-PET dataset}
  \label{fig:PET}
\end{minipage}
}

\section{Multi-Focus Image Fusion}
In order to verify the generalizability of the model, we tested it directly using 639 pairs of images from the RealMFF dataset as a training target for the multifocus task. Multi-focus image fusion was tested using 71 pairs of images from RealMFF, 20 pairs of images from Lytro to validate its effectiveness, and the fusion results were compared with state-of-the-art methods including DIFNet (2020) \cite{jung2020unsupervised}, CUNet (2020) \cite{deng2020deep}, SDNet (2021), U2Fusion (2020), DeFusion (2022) \cite{liang2022fusion}, ZMFF (2023) \cite{hu2023zmff}, FILM (2024) \cite{zhao2024image} and Ours.

The qualitative and quantitative results on the RealMFF dataset are shown in \textbf{Table~\ref{tab:lytro_realmff_results}}. In the multifocal image fusion experiments, qualitative comparisons are made by observing the fusion results generated by different algorithms.DIFNet and CUNet perform better in maintaining clarity and structural integrity, but are prone to some blurred areas when dealing with details in complex scenes.SDNet and DeFusion can better fuse different levels of details in multifocal images, especially in image depth perception and edge sharpness.U2Fusion, despite its improvement in detail recovery, may suffer from color shifts or texture distortion in some scenes.In contrast, the method in this paper effectively avoids the blurring or distortion that occurs in other methods and shows excellent fusion results.

When quantitatively evaluating the multifocal image fusion methods for the RealMFF dataset, the algorithms show significant differences in various metrics. First, in terms of entropy metrics, the Ours algorithm performs best, slightly higher than FILM, indicating that Ours is richer in terms of information content. In terms of standard deviation, Ours also scored the highest with 56.93, slightly higher than DeFusion, implying that the fused images generated by Ours have higher contrast and clarity. In terms of structural fidelity, SDNet still performs well and slightly outperforms Ours, but Ours is not inferior to other methods while maintaining structural integrity. In terms of visual information fidelity, FILM leads with a score of 1.52, followed by the proposed method, indicating its strong ability to preserve visual information in the fused images.


\begin{table*}[h!]
\centering
\captionsetup{font=small}
\caption{Quality evaluation on Lytro and RealMFF datasets.
The \bestbox{red} and \secondbox{blue} represent the best and second-best values.}
\label{tab:lytro_realmff_results}

\footnotesize
\setlength{\tabcolsep}{4pt}
\renewcommand{\arraystretch}{1.15}

\begin{tabular}{cc}

\begin{tabular}{lcccccc}
\multicolumn{7}{c}{\textbf{Lytro Dataset}} \\
\hline
Method & EN$\uparrow$ & SD$\uparrow$ & SF$\uparrow$ & VIF$\uparrow$ & Qabf$\uparrow$ & AG$\uparrow$ \\
\hline
DIFNet   & 7.43 & 52.52 & 11.47 & 0.73 & 0.54 & 4.30 \\
CUNet    & 7.25 & 45.78 & 15.54 & 0.71 & 0.65 & 5.58 \\
SDNet    & 7.47 & 55.25 & 16.88 & 0.84 & 0.69 & \Second{5.84} \\
U2Fusion & 7.30 & 51.95 & 14.83 & 0.83 & 0.65 & 5.60 \\
DeFusion & 7.52 & 56.65 & 11.55 & 0.80 & 0.55 & 4.35 \\
ZMFF     & 7.53 & 56.96 & \best{18.84} & 0.93 & \Second{0.70} & \best{6.76} \\
FILM     & \Second{7.56} & \Second{58.80} & \Second{15.61} & \best{1.13} & 0.67 & 5.46 \\
Ours     & \best{7.60} & \best{59.97} & 15.25 & \Second{1.11} & \best{0.76} & 5.63 \\
\hline
\end{tabular}

&

\begin{tabular}{lcccccc}
\multicolumn{7}{c}{\textbf{RealMFF Dataset}} \\
\hline
Method & EN$\uparrow$ & SD$\uparrow$ & SF$\uparrow$ & VIF$\uparrow$ & Qabf$\uparrow$ & AG$\uparrow$ \\
\hline
DIFNet   & 7.01 & 51.17 & 10.78 & 0.89 & 0.69 & 3.96 \\
CUNet    & 6.72 & 38.97 & 13.59 & 0.77 & 0.65 & 4.81 \\
SDNet    & 6.95 & 50.96 & \best{15.22} & 0.93 & 0.71 & 5.02 \\
U2Fusion & 6.77 & 48.49 & 14.07 & 0.95 & 0.70 & \Second{5.09} \\
DeFusion & 7.09 & \Second{56.42} & 11.24 & 0.98 & 0.69 & 4.08 \\
ZMFF     & 6.99 & 51.15 & 13.93 & 0.94 & 0.70 & 4.95 \\
FILM     & \Second{7.12} & 55.00 & 13.70 & \best{1.52} & \Second{0.75} & 4.80 \\
Ours     & \best{7.21} & \best{56.93} & \Second{14.12} & \Second{1.50} & \best{0.85} & \best{5.14} \\
\hline
\end{tabular}

\end{tabular}
\end{table*}

{
\vspace{12pt}  
\centering
\begin{minipage}{\linewidth}
  \centering
  \includegraphics[width=0.9\linewidth]{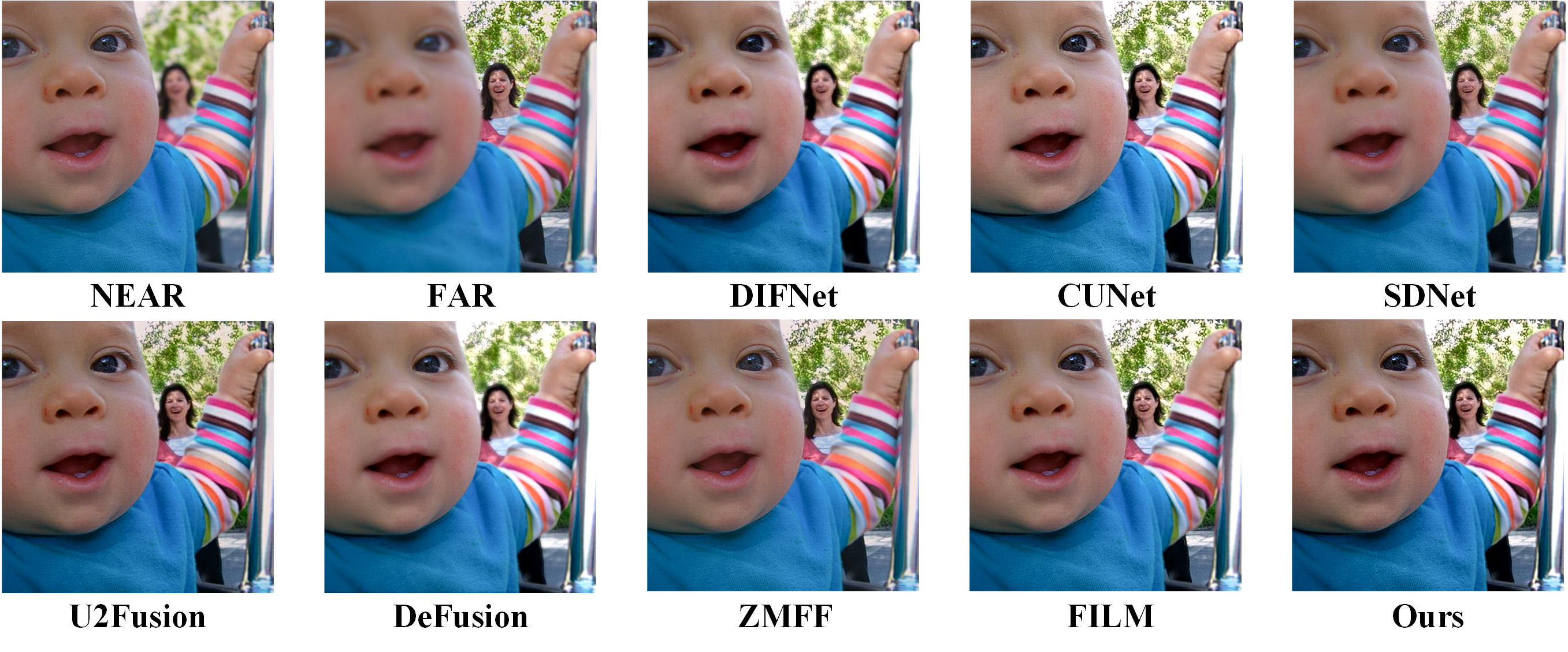}
  \captionof{figure}{Visual comparison of fusion results in multifocal image fusion tasks.}
  \label{fig8}
\end{minipage}
}

In the multi-focus image fusion experiments on Lytro dataset as shown \textbf{Table~\ref{tab:lytro_realmff_results}}, by observing the fusion results produced by different algorithms, it is found that DIFNet and CUNet perform better in some detail areas, especially in the focus area under the perspective of the human eye, but sometimes there is transition blurring or local distortion. The fusion results of SDNet and U2Fusion are more balanced and can effectively retain the structural information and details of the image, especially in depth perception and image edges. DeFusion and ZMFF are able to fuse the multi-focus images better, but in some complex scenes, the recovery of the details and the contrast is not perfect, which results in some areas not being clear enough.FILM in terms of detail recovery performance is better, but over-enhancement may occur in some high-contrast areas. Comparatively speaking, the fusion effects demonstrated in this chapter are more balanced and of higher quality in these challenging scenes, especially in terms of depth perception and natural transitions of focus shifts, which can provide a more realistic visual experience.

From the quantitative results, the methods in this chapter demonstrate excellent performance across several key metrics. In terms of entropy, the proposed method slightly outperforms FILM and ZMFF, indicating that the generated fused images contain richer information. For standard deviation, the proposed method achieves the highest score, suggesting superior contrast and sharpness. Regarding structural fidelity, although ZMFF shows marginally better performance, the proposed method maintains strong structure preservation and achieves a favorable balance between contrast enhancement and detail recovery. In terms of visual information fidelity, the proposed method ranks second only to FILM, reflecting a high level of fidelity preservation. Furthermore, with respect to average gradient in RealMFF dataset, the proposed method outperforms SDNet and U2Fusion, indicating enhanced image clarity and sharpness. Overall, the proposed method exhibits a well-balanced and competitive performance across multiple metrics, particularly excelling in clarity, contrast, information fidelity, and detail preservation, which demonstrates its effectiveness for multifocal image fusion.

\section{Object Detection}

To evaluate the proposed fusion method in high-level vision tasks, object detection experiments are conducted on the MSRS dataset using YOLOv5. Detection results from infrared images, visible images, and fused images are compared in terms of mAP@0.5 and mAP@0.5:0.95 for "Person" and "Car" categories, as shown in \textbf{Fig.~\ref{fig:leida}} and visualized in \textbf{Fig.~\ref{fig:object}}.

Fused images generally achieve better detection performance than single-modality inputs. For the "Person" category, visible images yield low mAP@0.5:0.95 due to complex backgrounds, whereas infrared images perform better. Our method further improves performance, achieving 0.930 in mAP@0.5 and 0.599 in mAP@0.5:0.95, indicating effective preservation of thermal targets with clear structural information.
For the "Car" category, the proposed method achieves 0.888 in mAP@0.5 and 0.695 in mAP@0.5:0.95, outperforming infrared input and most existing fusion methods. While some methods retain partial texture or intensity information, they often suffer from blurred boundaries or background noise. Our approach balances infrared saliency and visible details, enhancing detection confidence and localization accuracy.
Overall, our method reaches the highest average mAP@0.5 and mAP@0.5:0.95 across categories, demonstrating robust performance compared with methods like ReCoNet, LRRNet, TarDAL, BDLFusion, SwinFusion, CDDFuse, and TDFusion.

{
\vspace{12pt}  
\captionsetup[figure]{labelfont={bf},labelformat={default},labelsep=period,name={Fig.}}
\centering
\begin{minipage}{\linewidth}
  \centering
  \includegraphics[width=0.95\linewidth]{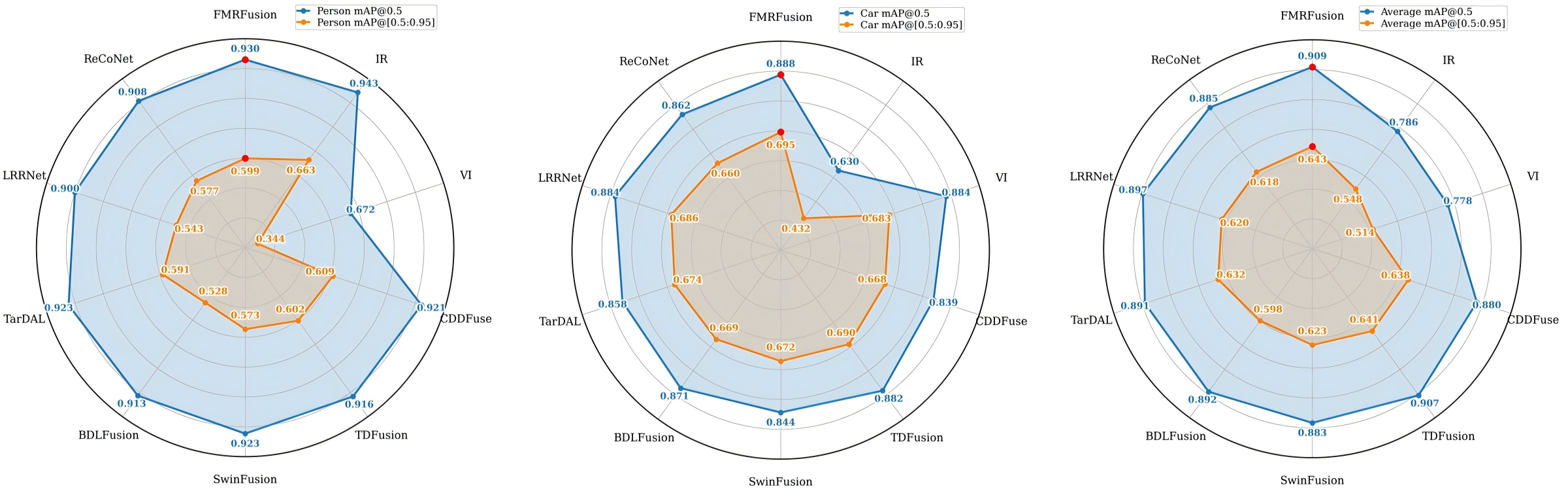}
  \captionof{figure}{Visual results of the object detection task on the MSRS dataset.}
  \label{fig:leida}
\end{minipage}
}

{
\vspace{12pt}  
\captionsetup[figure]{labelfont={bf},labelformat={default},labelsep=period,name={Fig.}}
\centering
\begin{minipage}{\linewidth}
  \centering
  \includegraphics[width=0.9\linewidth]{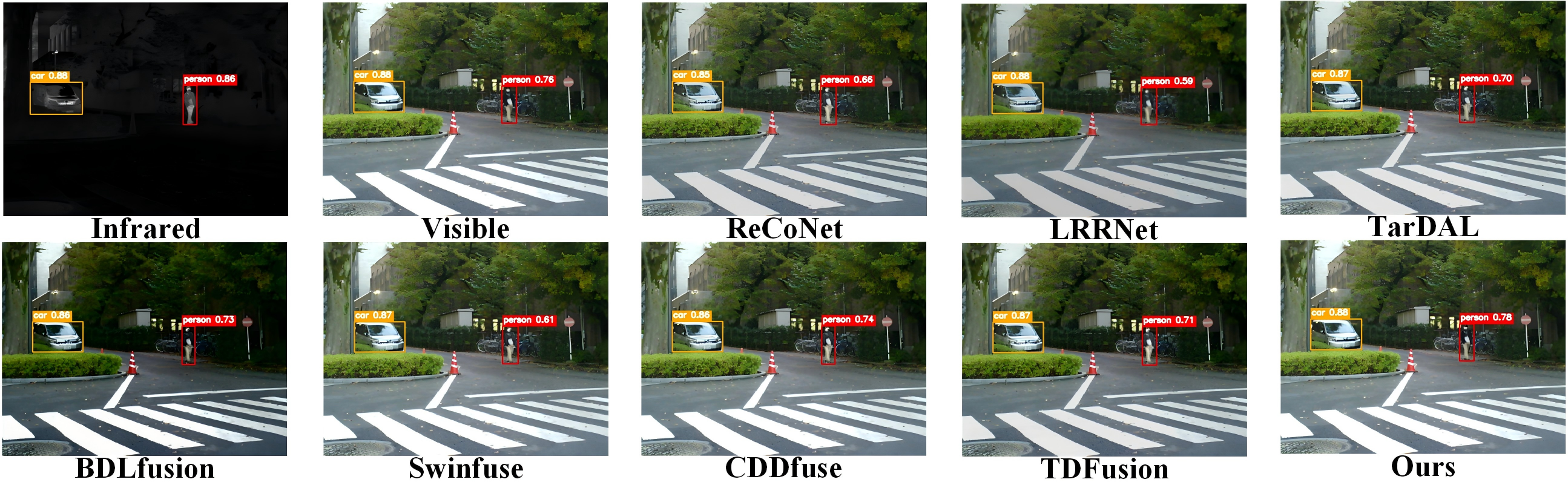}
  \captionof{figure}{Visual results of the object detection task on the MSRS dataset.}
  \label{fig:object}
\end{minipage}
}


\section{Conclusion}

In this work, we propose FMRFusion, a frequency-aware multi-view representation learning network that targets several critical challenges in heterogeneous data fusion. First, insufficient decoupling of view-specific information is addressed through explicit high-/low-frequency decomposition, which alleviates the loss of structural and intensity details commonly observed in conventional single-branch fusion architectures. Second, the imbalance between local detail preservation and global contextual modeling under degraded imaging conditions is mitigated by a hybrid representation learning strategy that integrates localized gradient-sensitive filtering with global dependency modeling, enabling robust structural consistency and fine-detail enhancement. Third, suboptimal cross-view information interaction is improved by incorporating a Cross-View Complementary Interaction Module, which explicitly facilitates complementary information exchange between radiative intensity cues and reflected structural information, supporting effective intensity enhancement and structure-aware denoising within a unified attention-driven framework. Furthermore, a Flow-Matching-based enhancement mechanism is introduced, which leverages the initial fusion result as a prior and learns a continuous transformation for progressive refinement, thereby improving saliency, contrast, and fine-detail representation.Extensive experiments demonstrate strong performance on standard infrared–visible fusion benchmarks and consistent cross-task generalization in medical image fusion and multi-focus image fusion, validating the effectiveness of the proposed approach for multi-view image fusion.

\section*{Authorship contribution statement}
\label{Authorship contribution statement}
Tao Zhou: Methodology, Formal analysis, Software, Writing – original draft. Yunlong Liu: Conceptualization, Data curation. Qinghui Chen: Validation, Software, Investigation. Zekai Zhang: Validation. Minlong Sun: Validation. Changlin Bian: Validation. Dagang Li: Validation. Wenmin Wang: Validation. Jinglin Zhang: Software, Review \& editing, Project administration, Funding acquisition.

\section*{Declaration of Competing Interest}
\label{Declaration of Competing Interest}
The authors declare that they have no known competing financial interests or personal relationships that could have appeared to influence the work reported in this paper.

\section*{Acknowledgments}
\label{Acknowledgments}
This work was supported in part by the joint funds of the National Natural Science Foundation of China under Grant U24A20221, Key R\&D Program of Shandong Province of China under Grant 2023CXGC010112, Distinguished Young Scholar of Shandong Province under Grant ZR2023JQ025, Taishan Scholars Program under Grant tstp20250708, Major Basic Research Projects of Shandong Province under Grant ZR2022ZD32.

\section*{Data availability}
\label{Data availability}
Data will be made available on request.

\bibliographystyle{elsarticle-num-names} 
\bibliography{cas-refs}
\end{document}